\documentclass{article}

\PassOptionsToPackage{numbers, compress}{natbib}
 \usepackage[preprint]{neurips_2026}

\usepackage[utf8]{inputenc} 
\usepackage[T1]{fontenc}    
\usepackage{url}            
\usepackage{booktabs}       
\usepackage{amsfonts}       
\usepackage{nicefrac}       
\usepackage{microtype}      
\usepackage{xcolor}         
\usepackage{graphicx,verbatim}
\usepackage{multirow}
\usepackage{multicol}
\usepackage{cite}
\usepackage[breaklinks]{hyperref} 
\usepackage{amssymb,amsmath}
\usepackage[most]{tcolorbox}
\newtcolorbox{promptbox}{
    colback=gray!3,
    boxrule=0.5pt,
    arc=2mm,
    left=4mm,
    right=4mm,
    top=2mm,
    bottom=2mm,
    enhanced
}

\title{Error Detection for PET/CT Radiology Reports: Domain-Specific vs Large Language Models}

\workshoptitle{GenAI4Health}

\author{Hermione Warr\textsuperscript{1} \quad Harry Anthony\textsuperscript{1} \quad Lilli Freischem\textsuperscript{2} \quad Yasin Ibrahim\textsuperscript{1} \\
\textbf{Daniel R. McGowan}\textsuperscript{3,4} \quad \textbf{Konstantinos Kamnitsas\textsuperscript{1}} \\
\textsuperscript{1}Department of Engineering Science, University of Oxford, UK \\
\texttt{\{first\_name.last\_name\}@eng.ox.ac.uk}\\
\textsuperscript{2}Department of Physics, University of Oxford, UK \\
\textsuperscript{3}Department of Oncology, University of Oxford, UK \\
\textsuperscript{4}Department of Medical Physics and Clinical Engineering, Oxford University Hospitals NHS, UK
}

\begin{document}

\maketitle

\begin{abstract}
Errors in radiology reports can adversely affect patient treatment, yet automated report quality assurance remains challenging because errors are often subtle and require domain expertise to detect. Although large language models (LLMs) have recently been proposed for radiology report verification, their ability to detect clinically meaningful errors beyond chest X-ray datasets remains under-explored. To this end, we present the first systematic evaluation of language models for PET/CT report error detection, comparing compact domain-specific models with SOTA open-weight LLMs. We collected 30,633 oncology FDG PET/CT reports from 23 radiologists over 10 years. 
We trained domain-specific BERT models to detect clinically motivated synthetic reporting errors and evaluated alongside zero-/few-shot Qwen3-32B, Gemma-3-27B and Llama-3.3-70B on a held-out benchmark of 11,500 reports. 
A 15M-parameter model achieved 94.4\% balanced accuracy with a 5.8\% false-positive rate, compared with 84.0\% for the strongest prompted LLM. Task-specific adaptation of Llama-3.3-70B closed this performance gap (94.4\%) but retained substantially greater computational requirements.
Our results suggest that domain-specific training matters more than model scale for PET/CT report error detection, supporting compact models as an accurate and computationally efficient approach to automated radiology report quality assurance.
Code \href{https://github.com/hermionewarr/Rad_Report_Error_Detection}{here}.

\end{abstract}

\section{Introduction}
\label{sec:intro}

Diagnostic errors remain a significant challenge in radiology, occurring in approximately 3--5\% of radiological interpretations \citep{Maskell_2019}. 
Clinicians rely on FDG PET/CT radiology reports - which consist of a detailed \emph{findings} section and a brief summarising \emph{conclusions} section - to guide cancer staging and assess response to treatment. 
Errors in reports can alter clinical management, 
which motivates automated methods for quality control \citep{Abacha_2025}.
Detecting such errors is challenging because many involve subtle semantic differences, such as distinguishing physiological uptake from malignancy or recognising changes in disease burden, rather than explicit textual contradictions \citep{Alotaibi_2020, Warr_2026}.

Recent work has explored Large Language Models (LLMs) for automated radiology report verification, but evaluation has largely been limited to chest X-ray (CXR) datasets 
\citep{Gertz_2024, Sun_2025, Kim2_2026, Min_2022} and a few on CTs \citep{ Kim_2025, Mayes_2026}. 
PET/CT reports present additional challenges due to the need to integrate metabolic and anatomical information and their substantially greater length (typically $>2\times$ that of CXR reports \citep{mimic_paper, Warr_2026}), raising questions about the generalisability of CXR report error detection approaches.
Furthermore, hospitals' privacy and governance requirements often preclude the use of cloud-hosted LLMs, motivating locally deployable solutions. Domain-specific language models (LMs) can be developed around site-specific processes and governance requirements, enabling simpler integration into clinical workflows \citep{Min_2022, Cho_2025, Belcak_small_llm_2025}. These factors raise a fundamental question: are billion-parameter LLMs necessary for PET/CT report quality control, or can compact domain-specific LMs better detect clinically meaningful inconsistencies?

To address this question, we collected a dataset of 30,633 whole body FDG PET/CT oncology reports. To compensate for the lack of radiology reports containing labelled errors, 
we introduce a synthetic error generator and construct a benchmark comprising 11,500 held-out reports with inserted clinically relevant inconsistencies (e.g. errors in disease severity or location). 
We pre-train domain-specific BERT models using masked language modelling and findings--conclusion consistency objectives before supervised fine-tuning, and compare performance against SOTA open-source LLMs.
Despite containing around 1000$\times$ fewer parameters, our domain-specific LMs consistently outperform all evaluated LLM baselines, achieving higher accuracy and substantially lower false-positive rates (FPR). 
Our contributions are as follows:
\begin{enumerate}
    \item We provide the first systematic comparison between state-of-the-art open-source LLMs and domain-specific language models for oncology PET/CT report error checking.
    \item We introduce a controlled synthetic benchmark spanning five clinically motivated PET/CT reporting errors and evaluate robustness to synthetic lexical artefacts, unseen perturbation terminology, prompt design, and model confidence.
    \item We show that compact models pre-trained on in-domain PET/CT reports substantially outperform much larger generic LLMs, with a 15M-parameter BERT achieving 94.4\% balanced accuracy and a 5.8\% false-positive rate.
\end{enumerate}

\section{Methods and Materials}
\label{sec:method}
\begin{figure}[h]
\centering
\includegraphics[width=0.99\textwidth]{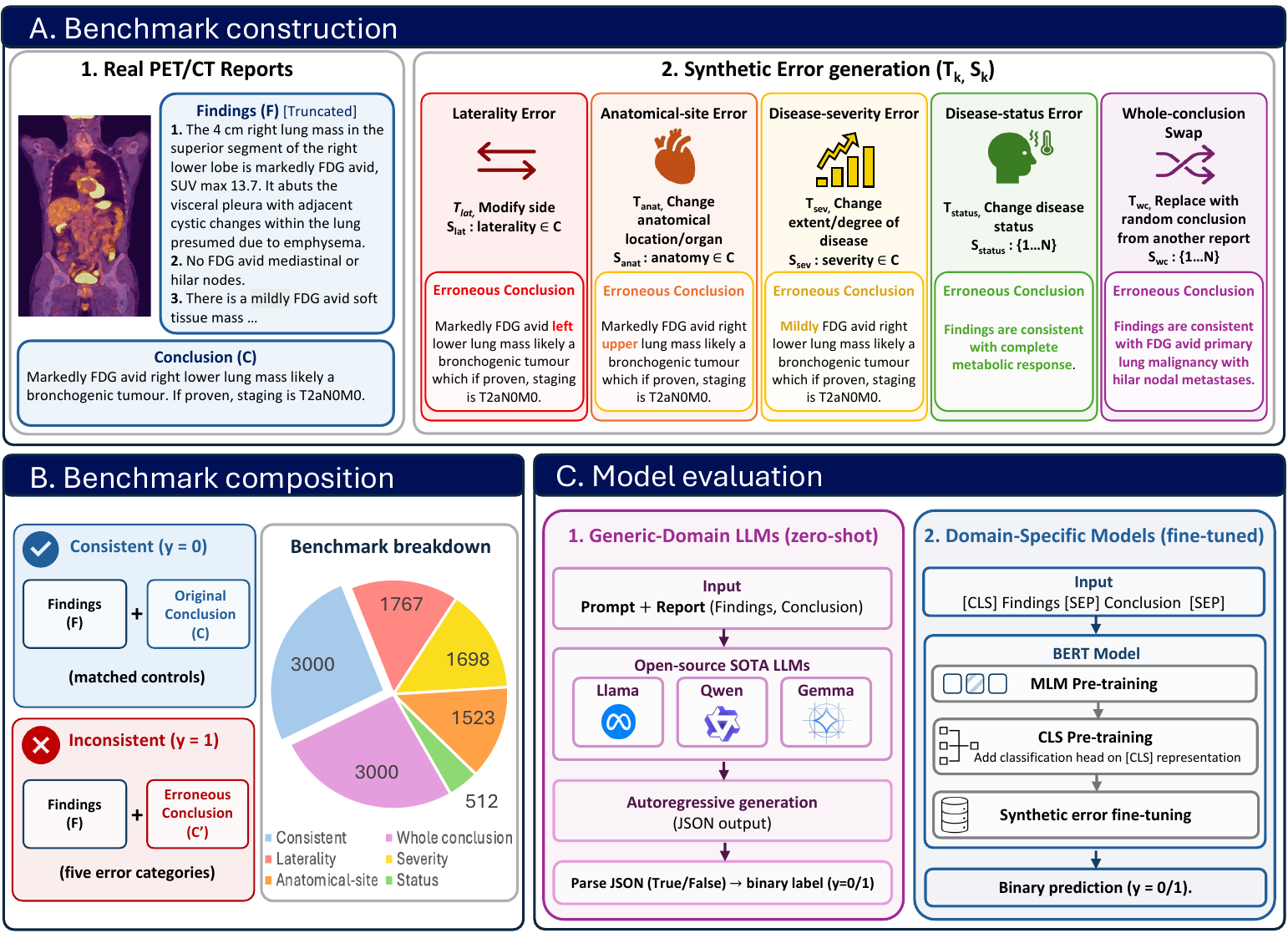}
\caption{Overview of the proposed PET/CT reporting error detection method. (A) Synthetic errors are generated from real reports by modifying only the conclusion section, ensuring all inserted errors are detectable from report text alone. Five clinically relevant error categories are considered: laterality, anatomical site, disease severity, disease status, whole conclusion mismatch. 
(B) Test-set composition. (C) Evaluation of prompted generic-domain LLMs and domain-specific models.
}
\label{fig:overview}
\end{figure}

\label{sec:meth_detection}

This work addresses the task of error detection in radiology reports from FDG PET/CT imaging. 
While not all reporting errors manifest as inconsistencies between the findings and conclusion sections, discrepancies between these sections are an important and actionable source of reporting error and are therefore the focus of this study.
Given a \emph{findings} section \(F\) and a \emph{conclusion} section \(C\), we represent a report as $r=(F,C)$. The objective is to determine whether $C$ is supported by $F$. 
We formulate this as a supervised text classification task, where a model maps each report \(r\) to a label \(y \in \mathcal{Y}\). The label space \(\mathcal{Y}\) depends on the experimental setting: in the binary setting, \(\mathcal{Y}=\{\text{consistent}, \text{inconsistent}\}\), whereas in the multi-class setting, \(\mathcal{Y}\) comprises one consistent class and five distinct error categories (Fig.~\ref{fig:overview}B). 

\subsection{Synthetic Error Generation Framework}
\label{sec:meth_err_gen}

Obtaining large collections of clinical report inconsistencies is challenging because reporting errors 
are not routinely annotated in clinical reporting systems \citep{Maskell_2019}. 
We therefore constructed a controlled synthetic consistency benchmark from real clinical FDG PET/CT reports.
For benchmark construction, original findings–conclusion pairs were treated as consistent controls, while synthetic inconsistencies were generated by modifying only the conclusion.
Let the set of original reports be $\mathcal{D}_{\mathrm{con}}=\{r_i\}_{i=1}^{N}$,
where each report is represented as $r_i=(F_i,C_i)$, with findings section $F_i$ and conclusion section $C_i$. 
Synthetic inconsistencies were generated using a family of perturbation operators
\begin{equation}
T_k: r_i \mapsto \hat{r}_i^{(k)},
\end{equation}
where each operator modifies only the conclusion section: $\hat{r}_i^{(k)}=(F_i,\hat{C}_i^{(k)})$.
The findings section remains unchanged, ensuring that any inconsistency arises solely from disagreement between the findings and conclusion. 
For each perturbation type $k$ (shown in Fig.~\ref{fig:overview}A), we define an eligibility set
\begin{equation}
S_k=\{i : T_k \text{ is applicable to } r_i\}.
\end{equation}
The resulting inconsistent dataset is therefore
\begin{equation}
\mathcal{D}_{\mathrm{inc}}=\bigcup_k \{\hat{r}_i^{(k)} : i \in S_k\},
\end{equation} 
with total size $|\mathcal{D}_{\mathrm{inc}}|=\sum_k |S_k|.$
The final benchmark is defined as
$\mathcal{D}=\mathcal{D}_{\mathrm{con}} \cup \mathcal{D}_{\mathrm{inc}},$
with labels:
\begin{equation}
y=
\begin{cases}
0, & r \in \mathcal{D}_{\mathrm{con}}, \\
1, & r \in \mathcal{D}_{\mathrm{inc}}.
\end{cases}
\end{equation}

We designed five clinically motivated perturbations to introduce controlled errors while preserving report fluency (Fig.~\ref{fig:overview}A), spanning lexical (e.g. laterality) and semantic errors requiring disease interpretation (e.g. severity or disease status).
The keyword-level perturbations ($T_{lat}, T_{anat}, T_{sev}$) were implemented using predefined antonym vocabularies (Appendix A), while whole-conclusion perturbations sampled replacement conclusions from either another report ($T_{wc}$) or the opposite disease-status class ($T_{status}$).
To evaluate performance on \emph{disease-status perturbations} ($T_{status}$), report conclusions were categorised as indicating either the presence of disease or no significant findings using a weakly supervised similarity-based labelling approach. 
Let $\phi(\cdot)$ denote an embedding function 
and $\mathcal{R}=\{R_j\}_{j=1}^{22}$ be a set of 22 canonical no-findings reference phrases (Appendix A).
The maximum similarity score is:
\begin{equation}
Sc(C)=\max_{R_j \in \mathcal{R}}
\cos\!\left(\phi(C),\,\phi(R_j)\right),
\end{equation}
where the score is given by cosine similarity.
Conclusions with $Sc(C)\geq0.8$ and no more than 15 words were labelled \emph{No Findings (NF)}; all remaining conclusions were labelled \emph{Disease}.
The resulting labels (NF, Disease) were used to sample replacement conclusions from the opposite disease-status class during synthetic benchmark generation. To assess label quality, a subset of reports was manually annotated. Verified labels were used for evaluation, while the automatically generated labels were used during training. 

\subsection{PET/CT radiology report dataset}
We collected a retrospective private dataset of whole-body FDG PET/CT radiology reports from oncology patients spanning a 10-year period at Oxford university hospitals (OUH). Only reports containing both findings and conclusion sections were included. The dataset was randomly split into training, validation, and test sets (approximately 80:10:10) at the patient level to prevent data leakage. Dataset statistics and split sizes are shown in Tab.~\ref{tab:dataset_stats}.

\textbf{Benchmark construction:}
%
Each of the 3,000 held-out test reports contributed one consistent example (the original report), and one instance of each applicable perturbation: 3{,}000 cross-pair mismatches, 1{,}767 laterality swaps, 1{,}698 severity flips, 1{,}523 site swaps, and 512 NF vs.\ disease swaps, yielding 11{,}500 evaluation reports (Fig.~\ref{fig:overview}B).

\begin{table}[t]
    \centering
    \caption{Dataset statistics for the FDG PET-CT dataset used in this study. Data collected between 2014-2024. Report length shows the tokens required for domain specific (DS) models are less than for generic-domain (gen) models.
    }
    \renewcommand{\arraystretch}{1.1} 
    \setlength{\tabcolsep}{5pt} 
    \resizebox{\textwidth}{!}{
    \begin{tabular}{lccc ccc ccc}
        \toprule
         \multirow{2}{*}{\textbf{Reports}} & \multirow{2}{*}{\textbf{Patients}}  
         & \multirow{2}{*}{\textbf{Radiologists}}
         & \multirow{2}{*}{\shortstack{\textbf{No. Unique} \\ \textbf{Words}}}
        & \multicolumn{3}{c}{\textbf{Report Length $\mu$ ($\sigma$)}} 
        & \multicolumn{3}{c}{\textbf{Data split}} \\ 
        \cmidrule(lr){5-7} \cmidrule(lr){8-10}
        & & & & \textbf{Words} & \textbf{Tokens DS} & \textbf{Tokens Gen.} &  \textbf{Train} & \textbf{Val} & \textbf{Test} \\
        \midrule
        30,633 & 20,193 & 23 & 14,663 & 184 (82) & 223 (100) & 286 (126)  & 24,525 & 3,108 & 3,000 \\
        \bottomrule
    \end{tabular}
    }
    \label{tab:dataset_stats}
\end{table}

\subsection{Error detection models}

All models were evaluated on the same findings--conclusion consistency task under two prediction settings: (i) binary error detection and (ii) multi-class error classification. Generic-domain LLMs were additionally evaluated for prompt sensitivity and confidence analysis. 
For confidence analysis, the models were prompted to return a confidence score from 1 to 5 alongside each prediction to see if confidence correlated with accuracy.
All experiments were conducted on a single NVIDIA RTX PRO 6000 GPU with 96\,GB of memory. 
Sequence length for all models was set so that 99.9\% of reports (and LLM prompts) would fit in context. 

\begin{figure}[h]
    \centering
    \includegraphics[width=\linewidth]{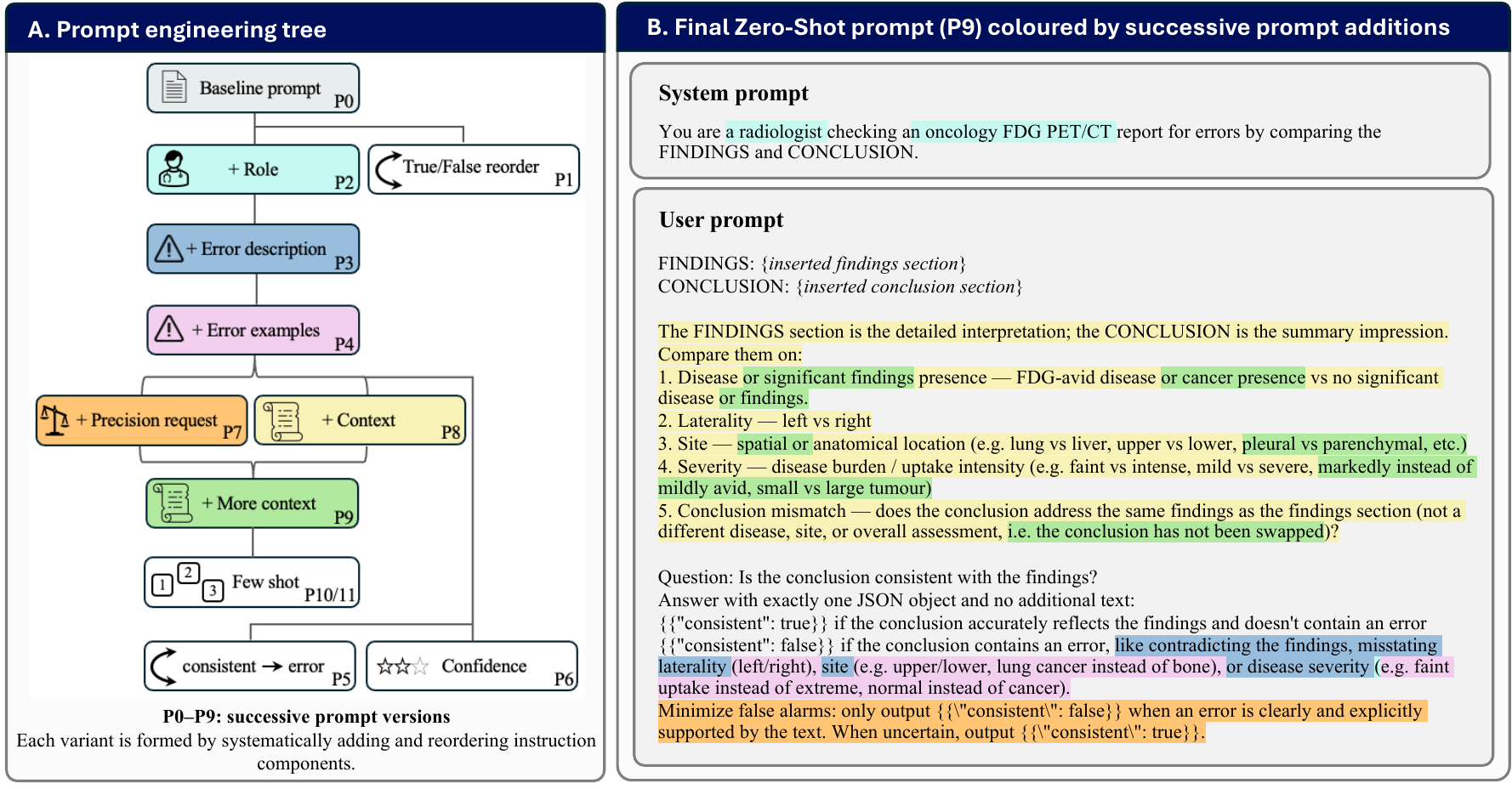}
    \caption{\textbf{Prompt engineering study} (A) The prompt tree corresponding to results in Tab.~3~(B,C,D). (B) System and user prompt provided to the LLMs (zero-shot cases). The un-highlighted text is baseline prompt - P0. Colours correspond to successive prompt additions (numbered in the prompt tree). The few-shot examples are added at the start of the user prompt.}
    \label{fig:prompt}
\end{figure}
\textbf{Generic-domain LLMs}
We compared three open-source LLMs: Qwen3-32B, Gemma3-27B, and Llama3-70B \citep{yang2025_qwen3, gemmateam_2025_gemma3, grattafiori_llama3_2024}. All models were evaluated locally using publicly available quantized checkpoints (Q6\_K GGUF for Qwen3 and Gemma 3; AWQ INT4 for Llama 3.3) under deterministic decoding (temperature=0). Thinking mode was disabled for Qwen3, as previous work found no benefit from  reasoning for text classification tasks \citep{Sprague_cot_2025}.
Each model received the \emph{findings} and \emph{conclusion} sections together with a \emph{prompt} asking whether the conclusion was consistent with the findings (Fig.~\ref{fig:overview}C). 
We systematically evaluated zero-shot prompt sensitivity using variants P0–P9, progressively varying label ordering (P1), expert-role assignment (P2), error descriptions/examples (P3–P4), task framing (P5), confidence elicitation (P6), false-positive minimisation (P7), and additional task context (P8–P9) (See Fig.~\ref{fig:prompt}). Our few-shot prompts P10\&11, additionally provided three and six (one of each class) fixed training-set examples. Full prompts are provided in Appendix A.


\textbf{Domain-specific models}
\label{sec:meth_DS_model} 
We evaluated four domain-specific models of
increasing capacity, based on a BERT style encoder backbone with a linear classification head \citep{bert_2019}.
The concatenated \emph{findings} and \emph{conclusion} sections were passed as a single input sequence, prefixed by a classification token (\texttt{[CLS]}), to the BERT encoder backbone. Let $h_{\mathrm{CLS}}=\phi(F,C)$ denote the contextual representation of the \texttt{[CLS]} token produced by the encoder backbone. Report consistency was then predicted by applying a linear classification head,
\begin{equation}
p(y\mid r)=\sigma\!\left(Wh_{\mathrm{CLS}}+b\right),
\end{equation}
where \(r=(F,C)\), \(W\) and \(b\) are learnable classifier parameters, \(\sigma(\cdot)\) is the sigmoid activation function for binary error detection ($y\in\{0,1\}$), or softmax for six-class classification ($y\in\{0,…,5\}$; consistent plus five error categories). 

All encoders were initialised with random weights and used a BPE tokenizer trained on the PET/CT training corpus (vocabulary size $\approx10$k) \citep{Warr_2026}.
The number of encoder layers, attention heads, and hidden dimension were scaled jointly, yielding four backbone configurations:
\begin{equation}
    (N_{layers}, N_{heads}, d_{model})\in\{(4,4,256),(6,6,368),(8,8,512),(12,12,768)\}
\end{equation}
denoted BERT$_4$, BERT$_6$, BERT$_8$, and BERT$_{12}$, respectively. All models used a max sequence length $S=600$ tokens and feed-forward dimension $d_{\text{ff}} = 4d_{\text{model}}$.

\textbf{Domain-specific training}
Model training consisted of three stages (Fig.~\ref{fig:overview}C) on $\sim$25k reports (Tab.~\ref{tab:dataset_stats}). 
First, the encoder backbone was pre-trained for 50 epochs using masked language modelling (MLM) (Masking probability = 0.15). Second, an intermediate consistency-aware objective was applied in which the model distinguished original findings--conclusion pairs from randomly mismatched pairs (20 epochs). This objective encourages the \texttt{[CLS]} representation to encode cross-section consistency before supervised learning. Finally, models were fine-tuned on the synthetic error detection task described in Sec~\ref{sec:meth_err_gen}. Models were optimised using AdamW, lr $2\times10^{-5}$, batch size 32 for 20 epochs. 
Pre-training details are provided in Appendix A.
During error-detection fine-tuning, errors were introduced to the training reports with probability 0.5, as described in Sec.~\ref{sec:meth_err_gen}. 
The models were trained to classify each report as consistent or inconsistent (binary), or into one of six classes: consistent, laterality, anatomical site, severity, disease status, or random conclusion mismatch (multiclass).

To improve robustness, we implemented several training strategies, including class-balanced sampling, balancing across error categories, and mitigation of synthetic lexical artefacts.
Laterality, site, and severity corruptions were less frequently occurring so were up-weighted (3:1 relative to the two replacement types). To reduce lexical cues introduced during severity perturbations, severity descriptors were sampled from three tiered synonym groups (mild, moderate, and severe tiers) while preserving grammatical agreement. In addition, within-tier synonym augmentation (e.g. intense$\rightarrow$extreme)  was applied to uncorrupted reports (with probability 0.4), reducing the likelihood that individual severity terms become reliable indicators of corrupted reports rather than findings--conclusion inconsistency. 
Error generator coverage, full synonym groups and perturbation vocabularies are provided in Appendix A.



\subsection{Evaluation and robustness experiments}

\textbf{Evaluation} We report accuracy, F1 score, true positive rate (TPR), false positive rate (FPR) and balanced accuracy. 
Because the five synthetic error categories occur at different frequencies in the test benchmark, balanced accuracy was defined as the average of macro-averaged recall across the five error categories and the true-negative rate (TNR; recall of consistent reports). 
We would also ideally like the probability of the model prediction to be calibrated:
\begin{equation}
    \mathbb{P}(\hat y = y |\hat
p = p) = p, ~~\forall~ p \in [0,1]
\end{equation} 
i.e. the predicted confidence $\hat p$ should match the empirical likelihood $p$ \citep{Guo_2017, Xiao_uncer_2022}.
One kind of measure of mis-calibration is the difference in expectation between confidence and accuracy, which can be approximated by ECE (expected calibration error) by partitioning N predictions into K equally-spaced bins and taking a weighted average of the bins’ accuracy/confidence difference which gives:
\begin{equation}
    ECE= \sum_{k=1}^K \frac{|B_k|}{N}|\text{acc}(B_k)-\text{conf}(B_k)|
\end{equation}
where $N$ is the number of samples and $B_k$ is the set of instances whose predicted probabilities $\hat p$ fall in bin $k$ - the interval $I_k=(\frac{(k-1)}{K},\frac{k}{K}]$. acc($B_k$) is the true fraction of positive instances in bin $k$ and conf($B_k$) is:
$\text{conf}(B_k) = \frac{1}{|B_k|}\sum_{i\in B_k} \hat p_i$. 
We also report the maximum calibration error (MCE):
\begin{equation}
    \mathrm{MCE}=\max_{m \in \{1,\ldots,K\}} \left|\operatorname{acc}(B_k)-\operatorname{conf}(B_k)\right|
\end{equation}
the largest absolute calibration gap across all bins.

\textbf{Severity term-holdout generalisation} 
Synthetic perturbations may introduce lexical cues that allow a classifier to associate individual words with the inconsistent class rather than detecting disagreement between findings and conclusion. We therefore evaluated generalisation to severity perturbations containing terms that were absent from synthetic-error training. Three BERT$_6$ and three BERT$_{12}$ models were retrained while withholding severity errors involving the term pairs \emph{mild/severe(ly)}, \emph{mild/intense(ly)} and \emph{faint/intense(ly)} respectively. Each model was evaluated separately on test errors involving its held-out terms and on the complementary set of severity errors. This experiment tests whether the model can identify inconsistencies involving severity terminology not previously encountered as a synthetic perturbation.


\section{Results} 
\begin{table}[b]
    \centering
    \caption{Error detection performance of generic-domain LLMs and domain-specific LMs. Domain-specific LMs over a minimum size (BERT$_6$) outperform general-domain LLMs on Bal. Acc. metric. Bal. Acc. here refers to the average of macro error recall and true negative rate (TNR / consistent class recall). Best results from bal. acc. prompt study (Tab.~\ref{tab:bert_prompt_var}) shown for LLMs for zero- and few-shot. Multi-class few-shot used one example per class.}
    \renewcommand{\arraystretch}{1}
    \setlength{\tabcolsep}{5pt}
    \resizebox{\textwidth}{!}{
    \begin{tabular}{ll c ccccc ccccc}
        \toprule
        & & & \multicolumn{5}{c}{\textbf{Binary Error Detection}} & \multicolumn{5}{c}{\textbf{Multi-class Error Classification}} \\
        \cmidrule(lr){4-8} \cmidrule(lr){9-13}
        \textbf{Domain} &\textbf{Model} &  \textbf{Param.}
        & \textbf{TPR}
        & \textbf{FPR}$\downarrow$
        & \textbf{Acc.}
        & \textbf{Bal. Acc.}
        & \textbf{F1}
        & \textbf{TPR}
        & \textbf{FPR}$\downarrow$
        & \textbf{Acc.}
        & \textbf{Bal. Acc.} 
        & \textbf{F1$_{macro}$}
        \\
        \midrule
        \multirow{8}{*}{Generic}&\textit{Zero-shot}\\
        &Gemma3 & 27B & 82.8 & 18.1 & 82.6 & 81.8 & 79.3 & 92.9 & 42.5 & 68.6 & 63.7 & 66.1 \\
        &Qwen3  & 32B &  86.9 & 18.6 & 85.4 & 82.8 & 82.1  & 87.4 & 16.4 & 61.8 & 71.9 & 59.0 \\
        &Llama3   & 70B & 84.6 & 14.2 & 84.9 & 84.0 & 82.0 & 87.1 & 20.8 & 67.4 & 73.3 & 65.2 \\
        \cmidrule(lr){2-13}
        &\textit{Few-shot}\\
        &Gemma3 & 27B & 65.6 & \textbf{4.9} & 73.3 & 79.5 & 71.7 & 82.5 & 14.7 & 70.9 & 74.4 & 67.0\\
        &Qwen3  & 32B &  83.4 & 14.3 & 84.0 & 83.3 & 81.1  & 79.5 & \textbf{6.2} & 57.0 & 73.0 & 52.7 \\
        &Llama3 & 70B & 84.0 & 15.4 & 84.2 & 83.6 & 81.2 &81.4 &8.7 &  67.1&76.7&63.7\\
        \midrule
        \multirow{4}{*}{Specific}&BERT$_4$  & 6.1M & 60.6  & 9.1  & 68.5 & 75.9 &  67.0 & 58.5 & 6.4 & 61.7 & 71.5 & 52.6 \\
        &BERT$_6$  & 15M & 94.7 & 5.8 & 94.6 & 94.4 & 93.1 & 96.1 & 7.4 & 90.0 & \textbf{90.3} & 86.9 \\
        &BERT$_8$  & 31M & 96.1 &  9.3  &  94.7 & 93.4  & 93.2 & 96.3 & 9.0 & 90.1 & 89.5 & 87.0 \\
        &BERT$_{12}$& 94M  &  \textbf{97.3} & 7.9 & \textbf{95.9} &  \textbf{94.6} & \textbf{94.7} & \textbf{97.3} & 8.2 & \textbf{91.5} & 90.2 & \textbf{88.4} \\
        \bottomrule
    \end{tabular}
    }
    \label{tab:results}
\end{table}
\begin{figure}[t]
\centering
\includegraphics[width=\textwidth]{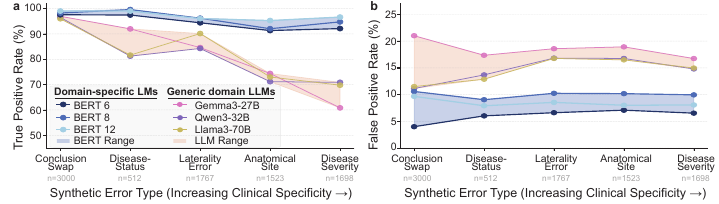}
\caption{TPR (a) and FPR (b) of domain-specific LMs (blue) and generic LLMs (red), for the binary detection task, across synthetic errors. LLMs have consistently lower TPR and higher FPR. 
Domain-specific models maintain high detection rates across error categories, whereas generic LLM performance decreases for several more semantically specific perturbations.
}
    \label{fig:clin_sig}
\end{figure}

\subsection{Generic vs Domain-specific:}
\begin{figure}[t]
\centering
\includegraphics[width=\textwidth]{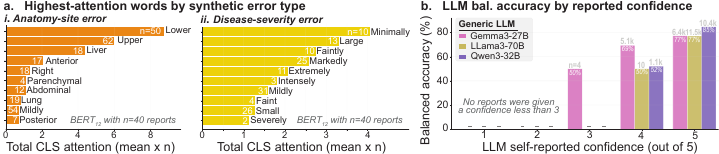}
\caption{(a) Words with a high attention score in BERT$_{12}$ from a random test subset of anatomy and severity errors. 
(b) LLM performance increases with self-reported confidence but, aside from Gemma, the LLMs mainly declare 5/5.
}
\label{fig:atn_conf}
\end{figure}

Tab.~\ref{tab:results} summarises performance on the held-out PET/CT error benchmark. 
Domain-specific models with six or more encoder layers substantially outperformed all generic zero- and few-shot LLM baselines despite containing orders of magnitude fewer parameters. BERT$_{12}$ achieved the highest binary balanced accuracy (94.6\%) compared with 84.0\% for the strongest LLM configuration (Llama3, P7), while BERT$_6$ achieved near-identical balanced accuracy (94.4\%) with only 15M parameters and the lowest false-positive rate among the domain-specific models (5.8\%).
This is particularly important for automated quality-assurance systems, where excessive false alarms may hinder clinical adoption. Performance therefore largely saturated beyond BERT$_6$, suggesting that increased model scale provided limited additional benefit once sufficient model capacity and in-domain pre-training were available.
The difference between model families was also consistent across synthetic error types (Fig.~\ref{fig:clin_sig}). Generic LLMs performed comparatively well on some natural domain linguistic inconsistencies but showed lower recall on more subtle site and severity perturbations, whereas BERT$_{6-12}$ maintained high recall across categories. Fig.~\ref{fig:atn_conf}a shows BERT$_{12}$ pays attention to expected key words.
The pre-training ablation (Tab.~\ref{tab:bert_prompt_var}A) indicates that the majority of the domain-specific models' performance derives from in-domain masked-language-model pre-training. For BERT$_6$, MLM pre-training increased balanced accuracy from 56.5\% to 93.7\%, while the subsequent consistency-aware pre-training objective increased it further to 94.4\%. The incremental benefit of this second objective diminished with model size, whereas MLM pre-training remained crucial across all evaluated capacities.


\begin{table}[t]
    \centering
    \caption{(A) Ablation study of domain-specific BERT pre-training. CLS token pre-training has less of an impact as model size increases but MLM pre-training remains crucial.
    (B, C, D) Prompt engineering study for each generic LLM. Best bal. acc. results in Tab.~\ref{tab:results}. Red indicates FPR>30\%.}
    \setlength{\tabcolsep}{4pt}
    \begin{minipage}[t]{0.515\textwidth}
        \centering
        \renewcommand{\arraystretch}{1.13}
        \resizebox{\linewidth}{!}{
        \begin{tabular}{llllll}
        \multicolumn{6}{c}{\large\bfseries A. BERT Pretraining Ablation Study} \\[0.3em]
            \toprule
            \textbf{Model} & \textbf{Pretrain} & \textbf{TPR}  & \textbf{FPR}$\downarrow$ 
            & \textbf{Bal. Acc.} & \textbf{F1} \\
            \midrule
            \multirow{3}{*}{BERT$_{4}$}  & from scratch   & 35.8 & 31.3 
            &  52.0 & 44.0 \\
            & with MLM & 55.3 \scriptsize(+19.5) & 13.3 \scriptsize(-18.0) 
            & 72.4 \scriptsize(+20.4) & 62.2 \scriptsize(+18.2) \\
            & with CLS &  60.6 \scriptsize(+5.3) & 9.1 \scriptsize(-4.2) 
            & 75.9 \scriptsize(+3.5) &  67.0 \scriptsize(+4.8)\\
            \midrule
            \multirow{3}{*}{BERT$_{6}$} & from scratch  & 35.2 & 21.5 
            & 56.5 & 46.3 \\
            & with MLM   & 93.5 \scriptsize(+58.3) & 6.0 \scriptsize(-15.5) 
            & 93.7 \scriptsize(+37.2) & 92.1 \scriptsize(+45.8) \\
            & with CLS & 94.7 \scriptsize(+1.2) & 5.8\scriptsize(-0.2) 
            & 94.4 \scriptsize(+0.7)& 93.1 \scriptsize(+1.0)\\
            \midrule
            \multirow{3}{*}{BERT$_{8}$} & from scratch & 26.0 & 15.9 
            & 55.2 & 41.1 \\
            & with MLM  & 95.7 \scriptsize(+69.7) & 9.5 \scriptsize(-6.4) 
            & 93.2 \scriptsize(+38.0) & 92.7 \scriptsize(+51.6) \\
            & with CLS &  96.1 \scriptsize(+0.4) &  9.3 \scriptsize(-0.4) 
            & 93.4 \scriptsize(+0.2) & 93.2 \scriptsize(+0.5) \\
            \midrule
            \multirow{3}{*}{BERT$_{12}$} & from scratch   & 100.0 &  100.0 
            & 50.0 & 42.5 \\
            & with MLM & 96.5 \scriptsize(-3.5) & 7.6 \scriptsize(-92.4) 
            & 94.5 \scriptsize(+44.5) & 94.1 \scriptsize(+51.6) \\
            & with CLS &  97.3 \scriptsize(+0.8) & 7.9 \scriptsize(+0.3) 
            &  94.6\scriptsize(+0.1) &  94.7\scriptsize(+0.6) \\
            \bottomrule
        \end{tabular}
        }
    \end{minipage}
    \hfill
    \begin{minipage}[t]{0.475\textwidth}
        \centering
        \small
        \renewcommand{\arraystretch}{1}
        \resizebox{\linewidth}{!}{
        \begin{tabular}{l cccc}
        \multicolumn{5}{c}{\large\bfseries B. Gemma3 Prompt Engineering Study} \\[0.3em]
            \toprule
            \textbf{Prompt} & \textbf{TPR} & \textbf{FPR$\downarrow$} & \textbf{Bal. Acc.} & \textbf{F1} \\
            \midrule
            P0 Baseline & 80.2 & 16.3 
            & 81.3 & 78.0 \\
            P1 True/False reorder & \textbf{96.8} & \textcolor{red}{71.1} & 
            62.8 & 79.1 \\
            P2 +Role & 85.6 & 22.9 & 
            80.9 &  \textbf{79.6} \\
            P3 +Error description  & 83.3 & 18.9 & 
            81.7 & 79.3 \\
            P4 +Error examples & 82.8 & 18.1 & 
            \textbf{81.8} & 79.3 \\
            P5 consistent$\rightarrow$ error & 84.4 & 25.9 
            & 78.9 & 77.6 \\
            P6 Confidence & 79.4 & 14.9 &  
            81.6 & 77.9 \\
            P7 +Minimise FP & 70.2 &  8.1 
            & 80.2 & 73.8 \\
            P8 +Error Context & 80.1 & 16.5 
            & 81.4 & 77.9 \\
            P9 +More Context & 70.6 & 8.8 & 80.4 & 73.9\\
            \midrule
            P10 Few shot (3-shot) &65.6 & 4.9 & 79.5 & 71.7 \\
            P11 Few shot (5-shot) & 58.8 & \textbf{3.1} & 76.9 & 67.7  \\
            \bottomrule
        \end{tabular}
        }
    \end{minipage}
    \begin{minipage}[t]{0.475\textwidth}
        \centering
        \small
        \resizebox{\linewidth}{!}{
        \begin{tabular}{l cccc}
        \\
        \multicolumn{5}{c}{\large\bfseries C. Qwen3 Prompt Engineering Study} \\[0.3em]
            \toprule
            \textbf{Prompt} & \textbf{TPR} & \textbf{FPR$\downarrow$} 
            & \textbf{Bal. Acc.} & \textbf{F1} \\
            \midrule
            P0 Baseline & 82.9 & 14.7 
            & 82.4 & 80.6 \\
            P1 True/False reorder  & \textbf{91.2} & \textcolor{red}{35.2} 
            & 77.1 & 78.9 \\ 
            P2 +Role & 86.8 & 20.4 
            & 82.0 & 81.5\\ 
            P3 +Error description & 87.5 & 20.9 
            & 82.2 & 81.8 \\ 
            P4 +Error examples & 87.9 & 20.8 
            & 82.5 & \textbf{82.1} \\ 
            P5 consistent$\rightarrow$ error & 86.9 & 19.1 
            & 82.7 & 82.0 \\
            P6 Confidence & 76.9 & 23.1 
            & 81.6 & 81.7 \\
            P7 +Minimise FP & 86.9 & 18.6 
            & 82.8 & \textbf{82.1} \\
            P8 +Error Context & 87.3 & 22.2 
            & 81.3 & 81.1 \\ 
            P9 +More Context & 88.8 & 25.0 & 80.9 & 81.2\\
            \midrule
            P10 Few shot (3-shot) & 83.4 & 14.3 & \textbf{83.3} & 81.1\\
            P11 Few shot (5-shot) & 77.8 & \textbf{8.3} & 83.1 & 79.1 \\
            \bottomrule
        \end{tabular}
        }
    \end{minipage}
    \hfill
    \begin{minipage}[t]{0.475\textwidth}
        \centering
        \small
        \resizebox{\linewidth}{!}{
        \begin{tabular}{l cccc}
        \\
        \multicolumn{5}{c}{\large\bfseries D. Llama3 Prompt Engineering Study} \\[0.3em]
            \toprule
            \textbf{Prompt} & \textbf{TPR} & \textbf{FPR$\downarrow$}  & \textbf{Bal. Acc.} & \textbf{F1} \\
            \midrule
            P0 Baseline & 90.8 & 28.0 & 80.6 &  81.6 \\
            P1 True/False reorder  & 87.9 & 25.0 & 80.5 & 80.5 \\
            P2 +Role & 93.8 & \textcolor{red}{37.4} & 77.6 & 80.1 \\
            P3 +Error description & 92.0  & 29.0  & 80.7 & 82.2 \\
            P4 +Error examples & 91.6  & 27.3 & 81.4 & \textbf{82.5} \\
            P5 consistent$\rightarrow$ error & 94.0 & \textcolor{red}{36.2}   & 78.1 & 80.7 \\
            P6 Confidence & \textbf{94.3} & \textcolor{red}{39.1} & 77.0 & 79.7 \\
            P7 +Minimise FP & 84.6 & 14.2 & \textbf{84.0} & 82.0 \\
            P8 +Error Context & 91.8 & \textcolor{red}{32.0} & 79.2 & 80.8 \\ 
            P9 +More Context & 85.4 & 16.2 & 83.7 & 81.9 \\
            \midrule
            P10 Few shot (3-shot) & 84.0 & 15.4 & 83.6 & 81.2\\
            P11 Few shot (5-shot) & 76.2 & \textbf{6.8} & 83.5 & 78.4 \\
            \bottomrule
        \end{tabular}
        }
    \end{minipage}
    \label{tab:bert_prompt_var}
\end{table}

\textbf{Fine-tuned LLM comparison} Since the generic LLM baselines above are evaluated under zero- and few-shot prompting, we additionally tested whether task-specific fine-tuning could close the performance gap. Parameter-efficient fine-tuning of LLama using LoRA achieved 94.4\% balanced accuracy, comparable to the domain-specific BERT models (93.4–94.6\%). However, this required adapting a substantially larger pretrained model: the LoRA configuration alone involved over 200M trainable parameters (compared with 15M parameters for the smallest working BERT) in addition to the frozen LLM backbone, and required substantially greater training time/resources. Thus, task-specific fine-tuning can recover the performance gap observed under prompting, but does not recover the computational efficiency of compact domain-specific models. See Appendix B for details.

\subsection{Robustness}
\textbf{Prompt engineering study} 
Tab.~\ref{tab:bert_prompt_var} shows performance differences were modest with increasing context (P0-P9), but were surprisingly sensitive to changes such as flipping the order of consistent true/false (P1) and asking the model to predict if the report contained an error true/false (P5) rather than consistency. 
Full prompt details (including multi-class prompt for Tab.~\ref{tab:results} results) can be found in Appendix B.
Few-shot prompting did not provide notable benefits to binary error detection - only Qwen's performance improved. However, few-shot prompting did improve multi-class predictions for all LLMs by reducing the FP rate in particular.

\textbf{Confidence and calibration}
We evaluated confidence in two complementary ways. First, generic LLMs were explicitly prompted to report confidence on a 1–5 scale. Balanced accuracy generally increased with reported confidence (Fig.~\ref{fig:atn_conf}b), but the models used only the upper end of the scale: no predictions received confidence below 3 and >99\% were assigned confidence 4 or 5. Self-reported confidence therefore indicated poor calibration for this task.
Second, we evaluated probabilistic calibration using model prediction probabilities. Expected calibration error (ECE) and maximum calibration error (MCE) were calculated using 10 equal-width probability bins (MCE filtered by n$\geq200$ to prevent sparsely populated bins dominating this metric)\citep{Guo_2017, Kadavath_2022}. Results show domain-specific models are better calibrated (lower ECE and MCE) than generic-LLMs (Fig.~\ref{fig:ece_mce}). Full calibration details provided in Appendix B.


\begin{figure}
    \centering
    \includegraphics[width=\linewidth]{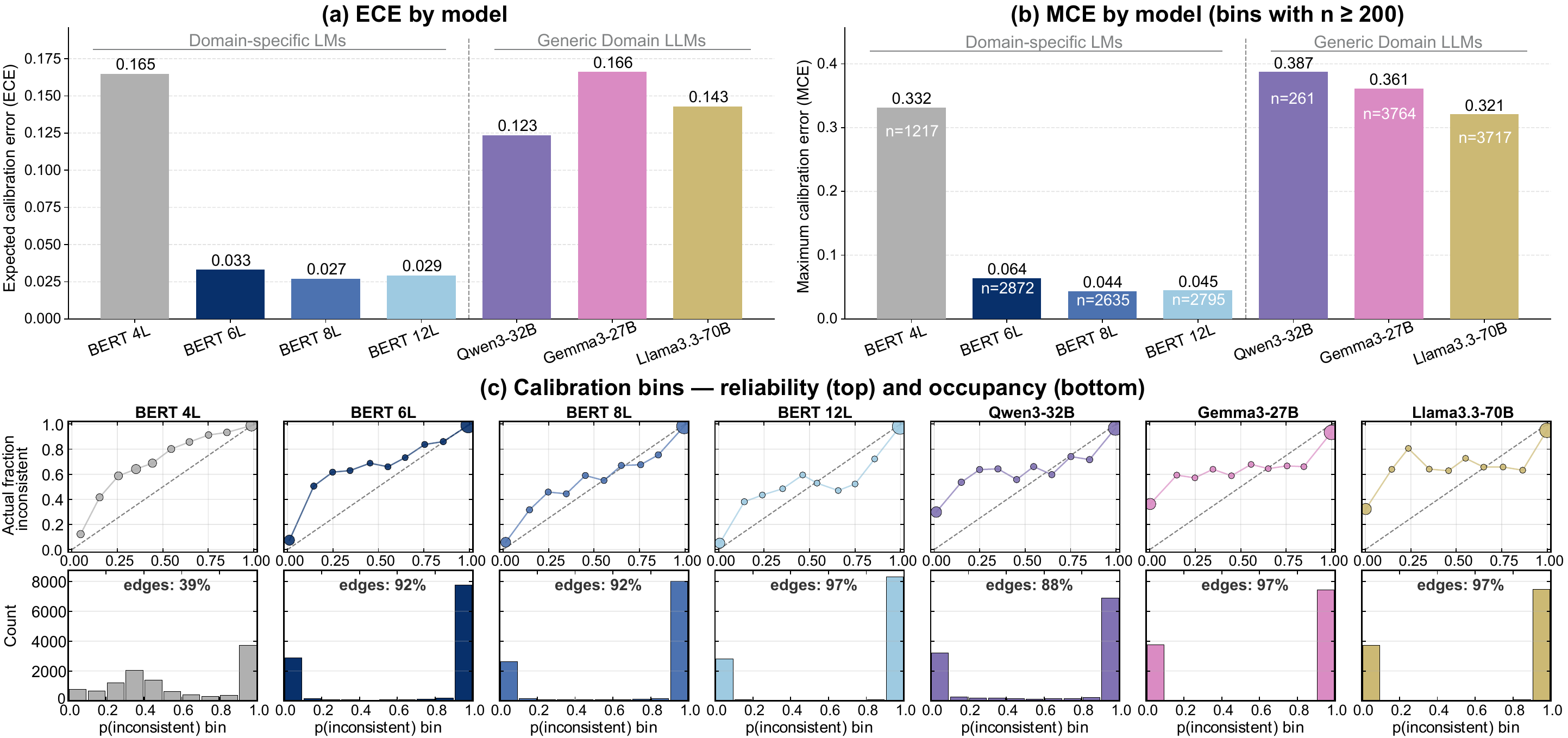}
    \caption{Calibration of \textit{p}(inconsistent). (a) Expected calibration error (ECE; 10 equal-width bins) on our synthetic benchmark (n=11,500). 
    The 4-layer BERT is shown in grey as a lower-capacity reference. (b) Maximum calibration error (MCE; same bins; $n\geq200$).
    (c) Reliability diagrams (above) and bin occupancy (below). Points show mean predicted probability vs empirical inconsistency rate within each bin (marker size = bin count); the dashed line is perfect calibration. Edge percentages are the share of predictions in the [0,0.1] and [0.9,1] bins. }
    \label{fig:ece_mce}
\end{figure}


\textbf{Generalisation to unseen errors} 
BERT$_{6}$ and BERT$_{12}$ retained substantial ability to detect severity errors involving term pairs withheld from synthetic-error training, although performance varied across pairs (see Appendix B). Recall on held-out errors decreased by 1.5–24.5 percentage points for BERT$_{6}$ and 8.8–25.9 for BERT$_{12}$. The largest reductions occurred when withholding mild(ly) perturbations, which comprise $\sim$40\% of severity-error training examples and also reduced performance on remaining severity errors. Withholding the less frequent faint(ly)/intense(ly) pair had substantially less effect, demonstrating partial generalisation to unseen perturbation terminology.

\textbf{Lexical null models} To test whether performance could be explained by superficial lexical properties or artifacts of the synthetic benchmark, we evaluated keyword-matching and bag-of-words overlap null models (Tab.\ref{tab:lexical_null}). 
\textit{Keyword} flags findings/conclusion discordance on laterality, site, and severity swap-term groups; \textit{Overlap} predicts inconsistent when bag-of-words Jaccard similarity falls below the validation-calibrated threshold ($0.07$); and \textit{Union} fires if either baseline does.
The strongest lexical baseline (union of both) achieved 64.2\% accuracy and 59.9\% inconsistent-report recall, compared with 95.9\% and 97.3\% for BERT$_{12}$, while also producing substantially more false positives (23.4\% vs. 7.9\%). 
Performance remained poor for the lexical baselines across individual perturbation types, suggesting that domain-specific models' performance cannot be explained by simple detection of perturbation keywords or findings–conclusion lexical overlap alone.


\begin{table}[h]
\centering
\small
\caption{Performance of lexical null models versus BERT$_{12}$ on the test benchmark.  \textbf{(a)} Overall and \textbf{(b)} error type performance of lexical null models versus BERT$_{12}$ on the test benchmark.
For error-inducing types in \textbf{(b)}, table values are detection recall.}
\begin{minipage}[t]{0.49\textwidth}
\centering
\textbf{(a) Overall performance}\\[1pt]
\resizebox{\linewidth}{!}{
\renewcommand{\arraystretch}{1.04}
\begin{tabular}{lcccc}
\toprule
\textbf{Model} & \textbf{Acc. } & \textbf{Bal. Acc.} & \textbf{Er.\ recall} & \textbf{ FPR} \\
\midrule
Keyword null   & 56.5 & 65.5 & 45.0 & 10.7 \\
Overlap null   & 37.5 & 56.1 & 20.1 & 13.2 \\
Union null     & 64.2 & 69.9 & 59.9 & 23.4 \\
BERT$_12$ (ref.) & 95.9 & 94.6 & 97.3 & 7.9 \\
\bottomrule
\end{tabular}}
\end{minipage}
\hfill
\begin{minipage}[t]{0.49\textwidth}
\centering
\textbf{(b) Recall by error type}\\[1pt]
\resizebox{\linewidth}{!}{
\begin{tabular}{l c cccc}
\toprule
\textbf{Error type} & \textbf{\emph{n}} & \textbf{Keyword} & \textbf{Overlap} & \textbf{Union} & \textbf{BERT$_12$} \\
\midrule
Laterality swap      & 1767 & 52.9 & 2.8  & 53.4 & 96.0 \\
Site swap            & 1523 & 64.6 & 4.4  & 65.9 & 95.3 \\
Severity flip        & 1698 & 41.8 & 4.7  & 45.0 & 96.6 \\
NF vs.\ disease swap & 512  & 11.1 & 77.3 & 86.5 & 99.0 \\
Cross-pair mismatch  & 3000 & 37.8 & 37.1 & 64.5 & 99.0 \\
Consistent           & 3000 & 89.3 & 86.8 & 76.6 & 92.1 \\
\bottomrule
\end{tabular}}
\end{minipage}
\label{tab:lexical_null}
\end{table}

\textbf{Documented clinical errors}
We identified seven test-set reports with subsequent clinician addenda documenting errors in the original report: four laterality errors, two textual errors (typos), and one omitted negation. The best domain-specific model detected 4/7 (all laterality errors), while Llama3 detected 4/7 (three laterality and one textual error); neither model detected the omitted-negation error. Although too small for quantitative evaluation, this opportunistic sample provides a preliminary check on naturally occurring errors beyond the synthetic benchmark.

\section{Conclusion}

We presented the first systematic comparison of compact domain-specific language models and open-weight LLMs for findings–conclusion error detection in oncology FDG PET/CT reports. On a controlled benchmark spanning five clinically motivated synthetic inconsistency types, compact domain-specific models substantially outperformed zero-/few-shot generic LLMs. Fine-tuning Llama-3.3-70B closed this performance gap, but at substantially greater computational cost. Performance largely saturated at modest model capacity: a 15M-parameter BERT achieved 94.4\% balanced accuracy with a 5.8\% false-positive rate, matching the accuracy of the task-adapted 70B LLM. Our results suggest that, for PET/CT report error detection, domain-specific training matters more than model scale and can substantially lower computational cost. Future work will validate these findings on larger collections of naturally occurring errors and external datasets.

\section{Acknowledgements}
HW and YI are supported by the EPSRC Centre for Doctoral Training in Health Data Science (EP/S02428X/1).
HA is supported by a scholarship via the EPSRC Doc-
toral Training Partnerships programme (EP/W524311/1,
EP/T517811/1). LF acknowledges funding from the NERC Doctoral Training Partnership in Environmental Research Grant (NE/S007474/1).
This research has been conducted using NHS data accessed via the Thames Valley and Surrey Secure Data Environment (TVS SDE), as part of the NHS Research Secure Data Environment Network. The NHS Research SDE Network is jointly funded by NHS England, the Department of Health and Social Care (DHSC) and the Department for Business, Energy and Industrial Strategy (BEIS). Data collection and processing within TVS SDE is also funded in part by the NIHR Oxford Biomedical Research Centre at Oxford University Hospitals NHS FT.

\bibliographystyle{plainnat}
\bibliography{bib}

@article{Alotaibi_2020, 
title={{Diagnostic errors in clinical FDG-PET/CT}}, 
volume={132}, 
journal={European Journal of Radiology}, 
publisher={Elsevier}, 
author={Alotaibi, Norah A. and others}, 
year={2020},
}

@article{Maskell_2019, 
title={Error in radiology—where are we now?}, 
volume={92},  
journal={The British Journal of Radiology}, 
author={Maskell, Giles}, 
year={2019},  
}

@inproceedings{Abacha_2025, 
title={{MEDEC: A Benchmark for Medical Error Detection and Correction in Clinical Notes}}, 
publisher={ACL}, 
author={Ben Abacha, Asma and others}, 
year={2025}, 
}

@inproceedings{Cho_2025, 
 title={{CREPE: Rapid Chest X-ray Report Evaluation by Predicting Multi-category Error Counts}}, 
 booktitle={Empirical Methods in NLP}, 
 publisher={ACL}, 
 author={Cho, Gihun and others}, 
 year={2025},  
}

@article{Gertz_2024, 
 title={{Potential of GPT-4 for Detecting Errors in Radiology Reports: Implications for Reporting Accuracy}}, volume={311}, 
 journal={Radiology}, author={Gertz, Roman Johannes and others}, 
 year={2024}
}

@article{Kim_2025, 
title={{Large-Scale Validation of the Feasibility of GPT-4 as a Proofreading Tool for Head CT Reports}}, 
volume={314}, 
journal={Radiology}, 
author={Kim, Songsoo and others}, year={2025}
}

@article{Kim2_2026, 
title={{Improving Radiology Report Error Detection Using a Multipass LLM: Framework Development and Validation}}, 
volume={14}, 
journal={JMIR Med. Inform.}, 
author={Kim, Songsoo and others}, year={2026}
}

@article{Mayes_2026, 
 title={{Improving radiology reporting accuracy: use of GPT-4 to reduce errors in reports}}, 
 volume={51},
 journal={Abdominal Radiology}, 
 author={Mayes, Connor J. and others}, 
 year={2026},
}

@inproceedings{Min_2022,  
 title={{RRED: A Radiology Report Error Detector based on Deep Learning Framework}},  
 booktitle={Proceedings of the 4th Clinical NLP Workshop}, 
 publisher={ACL}, 
 author={Min, Dabin and others}, 
 year={2022},  
}

@article{Sun_2025, 
 title={Generative Large Language Models Trained for Detecting Errors in Radiology Reports}, 
 journal = {Radiology},
 volume = {325},
 number = {2},
publisher={RSNA}, 
author={Sun, Cong and others}, 
year={2025}
}

@inproceedings{Xiao_uncer_2022, 
address={Abu Dhabi, United Arab Emirates}, 
 title={Uncertainty Quantification with Pre-trained Language Models: A Large-Scale Empirical Analysis},
 booktitle={Findings of the Association for Computational Linguistics: EMNLP 2022}, 
 author={Xiao, Yuxin and others}, 
 year={2022},
 pages={7273–7284}
}

@inproceedings{Guo_2017, 
title={On Calibration of Modern Neural Networks},  
booktitle={Proceedings of the 34th International Conference on Machine Learning}, 
publisher={PMLR}, 
author={Guo, Chuan and others}, year={2017}, 
month=july, 
pages={1321–1330}, 
}

@article{Kadavath_2022, 
 title={Language Models (Mostly) Know What They Know}, 
  author={Kadavath, Saurav and others}, 
  year={2022}
}

@inproceedings{bert_2019,
  title={{BERT: Pre-training of Deep Bidirectional Transformers for Language Understanding}},
  author={Jacob Devlin and others},
  booktitle={ACL},
  year={2019},
}

@misc{grattafiori_llama3_2024,
      title={The Llama 3 Herd of Models}, 
      author={Aaron Grattafiori and others},
      year={2024},
      archivePrefix={arXiv},
      publisher = {Meta},
}

@misc{yang2025_qwen3,
      title={Qwen3 Technical Report}, 
      author={An Yang and others},
      year={2025},
      archivePrefix={arXiv},
}

@misc{gemmateam_2025_gemma3,
      title={Gemma 3 Technical Report}, 
      author={Gemma Team},
      year={2025},
      archivePrefix={arXiv},
}

@article{mimic_paper,
	title = {{MIMIC-CXR}, a de-identified publicly available database of chest radiographs with free-text reports},
	volume = {6},
	journal = {Sci Data},
    author = {Johnson, A.E.W. and others},
	year = {2019},
}

@inproceedings{Warr_2026, 
 title={{Specialised or Generic? Tokenization Choices for Radiology Language Models}}, 
 booktitle={ELAMI}, 
 publisher={Springer Nature Switzerland}, 
 author={Warr, Hermione and others}, 
 year={2026}
}

@inproceedings{Sprague_cot_2025,
    author ={Sprague, Zayne and others},
    title = {{To CoT or not to CoT? Chain-of-thought helps mainly on math and symbolic reasoning}},
    booktitle ={ICLR},
    year = 2025
}

@inproceedings{Dettmers_qlora_2023, 
title={QLORA: efficient finetuning of quantized LLMs}, booktitle={Proceedings of the 37th International Conference on Neural Information Processing Systems}, publisher={Curran Associates Inc.}, 
author={Dettmers, Tim and others}, 
year={2023}, pages={10088–10115}
}

@article{lora_2021, 
title={LoRA: Low-Rank Adaptation of Large Language Models}, 
DOI={10.48550/arXiv.2106.09685}, 
publisher={arXiv}, 
author={Hu, Edward J. and others}, 
year={2021} }

@article{Belcak_small_llm_2025, 
 title={{Small Language Models are the Future of Agentic AI}}, 
 DOI={10.48550/arXiv.2506.02153}, 
 publisher={arXiv}, 
 author={Belcak, Peter and others}, 
 year={2025}
 }

\newpage

\appendix
\section{Appendix: Materials and methods}
\label{sec:meth_detection}

\subsection{Synthetic Error Generation: Perturbation type breakdown}
\label{sec:append_meth_err_gen}

Defined here in more detail are the five clinically motivated perturbation operators that were considered to give us the five error types in main paper. 

\textbf{Laterality perturbation.}
For conclusions containing the laterality descriptors \textit{left} or \textit{right}, the perturbation swaps the descriptor with its opposite, producing
\begin{equation}
\hat{C}_i^{(\mathrm{lat})}=T_{\mathrm{lat}}(C_i).
\end{equation}

\textbf{Anatomical-site perturbation.}
Let $V_{anat}=\{a_1, ... , a_m \}$ denote a predefined vocabulary of anatomical locations. If a conclusion contains an anatomical term $a_j\in V_{anat}$, it is replaced with an antonym $a_\ell \in V_{anat}$, where $\ell\neq j$.
\begin{equation}
    \hat{C}_i^{(\mathrm{anat})}=T_{\mathrm{anat}}(C_i).
\end{equation}

The anatomical site keyword antonym pairs are $V_{anat}$=\{\{upper, lower\}, \{top, bottom\}, \{thoracic, abdominal\}, \{apical, basal\}, \{proximal, distal\}, \{anterior, posterior\}, \{lung, liver\}, \{lung, bone\}, \{pleural, parenchymal\}\}. Swaps are within pairs. 

\textbf{Disease-severity perturbation.}
Perturbations replace severity terms according to manually defined mappings that alter the reported disease burden while preserving grammatical structure. 
Severity flips combine keyword swaps (small $\leftrightarrow$ large; low-grade $\leftrightarrow$ high-grade) with three qualitative tiers (mild/faint/minimal; moderate; severe/marked/intense/extreme), replacing terms within or across tiers with morphology-matched synonyms (e.g.\ \textit{mildly}$\rightarrow$\textit{markedly}). Within tier swaps creates an augmentation (e.g. \textit{faint}$\rightarrow$\textit{minimal}) and between tiers creates an error. We applied the augmentation (within tier swap) with probability $p=0.4$ on reports containing those terms during training.
Key-word replacements correct preceding \textit{a/an} where needed.
\begin{equation}
    \hat C_i^{(sev)}=T_{sev}(C_i)
\end{equation}

\begin{figure}
    \centering
    \includegraphics[width=0.9\linewidth]{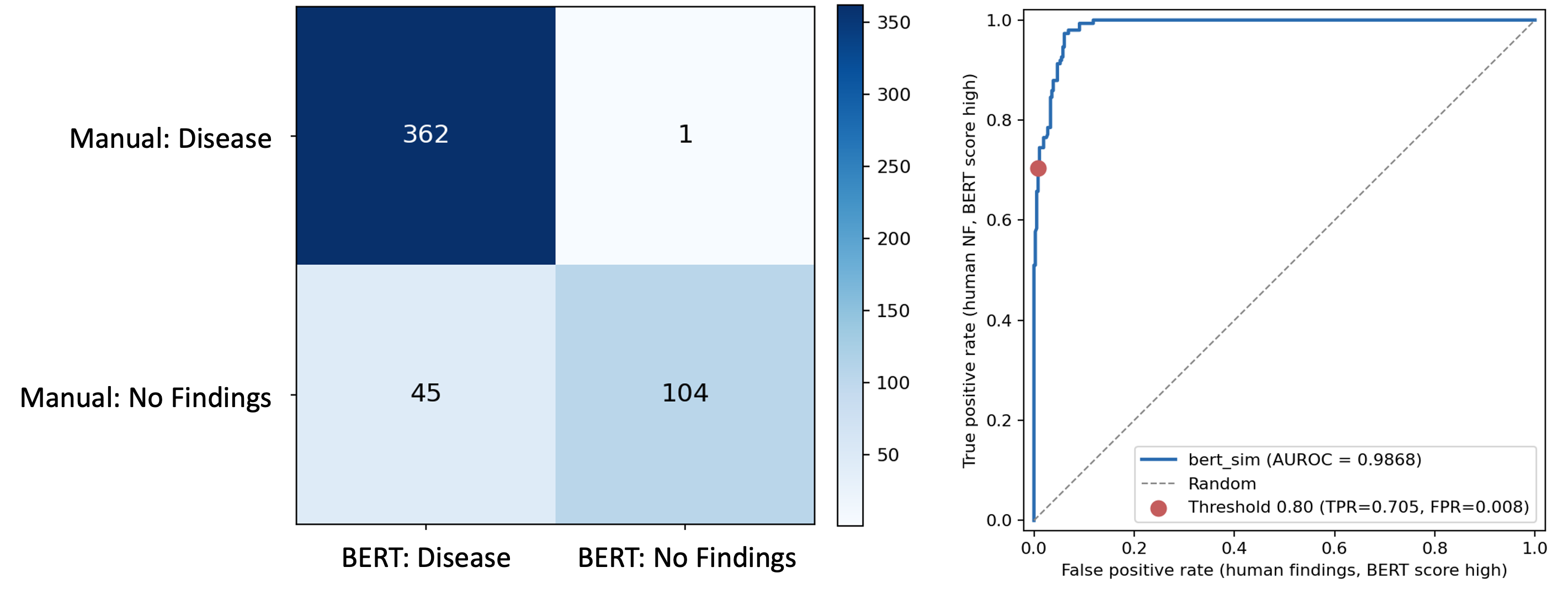}
    \caption{Results of Disease Status labeller. Acc 91.0\% with our threshold of 0.8 on the benchmark subset of 512 (548 were manually labelled but 35 were excluded as being indefinite). This manually labelled subset was used for evaluation but the labeller was used for generating labels for training.}
    \label{fig:nf_labels}
\end{figure}

\textbf{Disease-status perturbation.}
Disease-status perturbations were designed to swap conclusions describing active disease with conclusions describing no evidence of active disease and vice versa:
\begin{equation}
    \hat{r}_i^{(\mathrm{status})}=(F_i,C_j)
\end{equation}
where $i\neq j$.
Conclusions were assigned an no findings (NF) label based on maximum cosine similarity to 22 canonical no-findings reference phrases (similarity $\geq0.80$ and $\leq15$ words), excluding long patient-specific qualified statements. The threshold was set to ensure conclusions labelled NF were likely to be correct (Fig~\ref{fig:nf_labels} (left)). To evaluate the reliability of the automatically generated labels, a subset of reports was independently reviewed by a human to produce gold-standard NF/disease annotations. A stratified subset of 512 test conclusions, enriched for predicted NF cases, borderline similarities (0.75--0.90), and random disease examples, was manually reviewed, with indeterminate cases excluded. The weak labeller achieved 91\% agreement with reference annotations on definitive labels (AUROC 0.987) (Fig.~\ref{fig:nf_labels}).  

The 22 canonical no findings references phrases are listed below:
\begin{quote}
    No FDG avid disease demonstrated.\\
    No evidence of FDG-avid disease.\\
    No FDG avid residual/recurrent disease demonstrated.\\
    No FDG avid disease.\\
    No FDG-avid disease.\\
    No FDG avid recurrent disease demonstrated.\\
    No FDG-avid recurrence demonstrated.\\
    No FDG avid malignancy identified.\\
    Normal study. No FDG avid malignancy demonstrated.\\
    No evidence of FDG avid residual or recurrent disease.\\
    No PET-CT evidence of disease.\\
    No PET CT evidence of disease.\\
    No PET-CT evidence of malignancy.\\
    There is no evidence of FDG avid disease.\\
    No avid local, nodal or distant disease demonstrated.\\
    No metabolic evidence of recurrent disease.\\
    No abnormal FDG uptake demonstrated.\\
    No FDG avid recurrence or residual disease demonstrated.\\
    No FDG avid relapse.\\
    No FDG avid malignancy.\\
    No FDG avid abnormality demonstrated.\\
    No FDG avid systemic disease.
\end{quote}

\textbf{Whole-conclusion mismatch perturbation.}
For report $r_i$ a mismatched conclusion was obtained by replacing $C_i$ with the conclusion from another report $C_j$ $(j \neq i)$ similarly to disease status perturbations:
\begin{equation}
\hat{r}_i^{wc}=(F_i,C_j).
\end{equation}
This perturbation preserves the linguistic characteristics of authentic clinical conclusions while introducing report-level inconsistency.

These 5 perturbations are summarised in Tab.~\ref{tab:perturbations}.
\begin{table}[h]
\centering
\caption{Perturbation types used to introduce controlled inconsistencies between findings and conclusions.}
\label{tab:perturbations}
\renewcommand{\arraystretch}{1.1}
\resizebox{\linewidth}{!}{
\setlength{\tabcolsep}{8pt}
\begin{tabular}{l l l l}
\toprule
\textbf{Perturbation $T_K$} & \textbf{Eligibility $S_{k}$} & \textbf{Modification} & \textbf{Intended inconsistency} \\
\midrule
Laterality, $T_{lat}$&
Laterality $\in C$, $S_{lat}$ &
Swap \textit{left} $\leftrightarrow$ \textit{right} &
Incorrect laterality \\

Anatomical site, $T_{anat}$&
Anatomical location $\in C$, $S_{anat}$ &
Replace location &
Incorrect disease location \\

Disease severity, $T_{sev}$&
Severity descriptor $\in C$, $S_{sev}$&
Replace severity&
Incorrect disease burden \\

Disease status, $T_{status}$ & 
All reports, $S_{status} = \{1, \dots, N\}$&
Replace with opposite status conclusion 
& Incorrect disease status \\

Whole conclusion, $T_{wc}$ &
All reports, $S_{wc}=\{1, \dots, N\}$ &
Replace with another report's conclusion &
Global mismatch \\
\bottomrule
\end{tabular}
}
\end{table}

\subsection{Error generator coverage} 
\begin{figure}
    \centering
    \includegraphics[width=\linewidth]{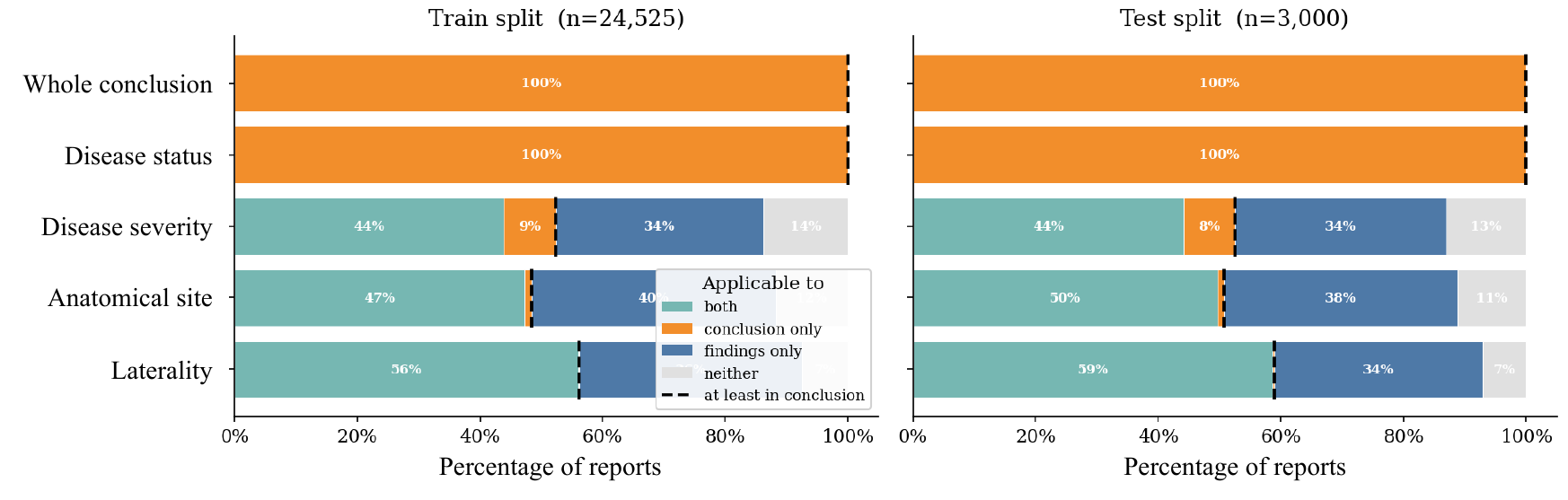}
    \caption{Error generator coverage across the train and test benchmark PET/CT corpus. Not all reports contained the error key-words that we were considering. }
    \label{fig:gen_cov}
\end{figure}

\begin{figure}
    \centering
    \includegraphics[width=\linewidth]{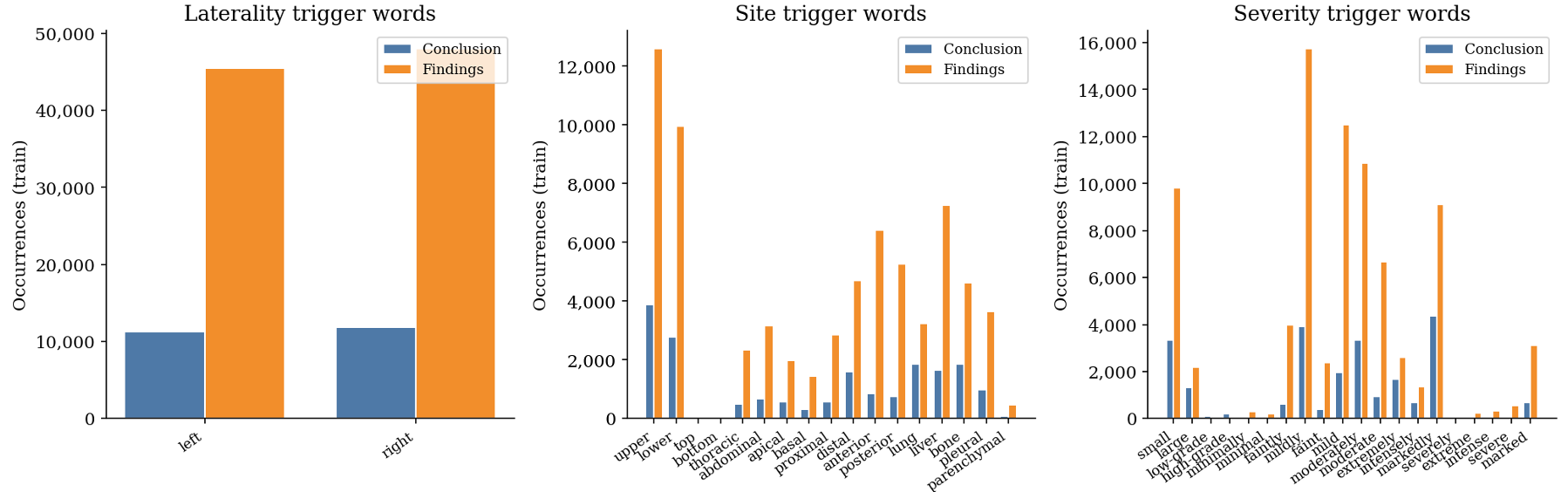}
    \caption{Key-word frequency in training corpus for laterality, anatomical site and severity error categories respectively. }
    \label{fig:trigger words}
\end{figure}

\begin{figure}
    \centering
    \includegraphics[width=\linewidth]{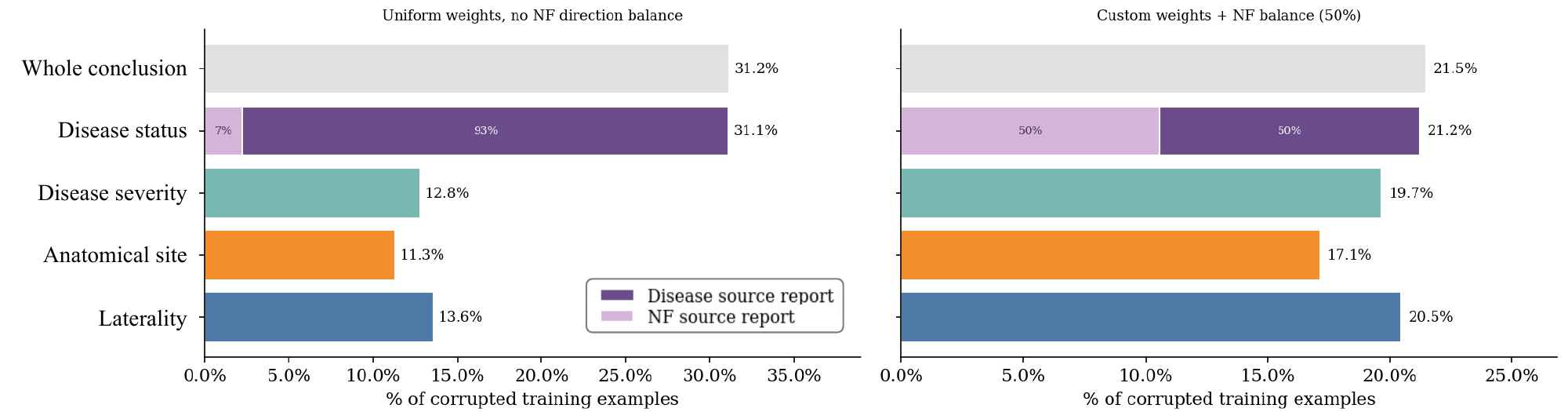}
    \caption{Class balancing effect on training distribution. Key-word swap errors were oversampled in training as not all reports contained these terms. Reports with disease status No findings (NF) were also oversampled to give better disease status training.}
    \label{fig:class_bal}
\end{figure}

On the training corpus, laterality, site, and severity generators were applicable to 56\%, 48\%, and 52\% of conclusions, respectively; NF vs.\ disease and cross-pair generators applied to all reports (Fig.\ref{fig:gen_cov})~. Applicable conclusions contained a median of one key-word (mean 1.6--1.7). The most frequent key-words were \texttt{left}/\texttt{right}, \texttt{upper}/\texttt{lower}, and \texttt{markedly}/\texttt{mildly} (see Fig.~\ref{fig:trigger words}). NF conclusions comprised 4.4\% of training reports.
To address some of these imbalances we oversampled under represented types in training (see Fig.~\ref{fig:class_bal}). 


\newpage
\subsection{Prompt engineering}

Here we provide some of the key prompts from the paper. Main Fig.~\ref{fig:prompt} details the successive prompt build-up.  P0-9 are zero-shot binary. P10,11 are Few shot. Multi-class is zero-shot and 6-shot (i.e. one of each class).

\textbf{P0 - Baseline prompt}\\
This is the non highlighted text in Fig.~\ref{fig:prompt} and all other prompts added to this text (excluding the multi-class prompt).
\begin{promptbox}
\footnotesize
\textbf{System prompt}\\
You are checking a report for errors by comparing the FINDINGS and CONCLUSION.
\vspace{0.5em}

\textbf{User prompt template}

FINDINGS:
\{findings\}

CONCLUSION:
\{conclusion\}

Question: Is the conclusion consistent with the findings?

Answer with exactly one JSON object and no additional text:
\verb|{"consistent": true}| 
if the conclusion accurately reflects the findings and doesn't contain an error.\\
\verb|{"consistent": false}| 
if the conclusion contains an error.
\end{promptbox}

\textbf{P1 - true/false order swap}\\
Here we swap the order of \texttt{\{"consistent":false\}, \{"consistent":true\}} and the text explaining them. The words used in the prompt stay the same only the order changes. 
\begin{promptbox}
\footnotesize
\textbf{System prompt}\\
You are checking a report for errors by comparing the FINDINGS and CONCLUSION.
\vspace{0.5em}

\textbf{User prompt template}

FINDINGS:
\{findings\}

CONCLUSION:
\{conclusion\}

Question: Is the conclusion consistent with the findings?

Answer with exactly one JSON object and no additional text:
\vspace{-2mm}
\begin{tcolorbox}[
    colback=red!8,
    colframe=red!30,
    boxrule=0pt,
    arc=1mm,
    left=0mm,
    right=0mm,
    top=0mm,
    bottom=0mm
]\texttt{\{"consistent": false\}}
if the conclusion contains an error.\\
\texttt{\{"consistent": true\}} 
if the conclusion accurately reflects the findings and doesn't contain an error.
\end{tcolorbox}
\end{promptbox}

\textbf{P4 - Error examples} \\
Second prompt tree branch point. Pink in fig.~\ref{fig:prompt}. This tells the model that it has a  radiologist role and that the kind of reports it is checking (added in P2 - light blue) as well as the error types (P3 - dark blue) and a few error examples.
\begin{promptbox}
\footnotesize
\textbf{System prompt}\\
You are a radiologist checking an oncology FDG PET/CT report for errors by comparing the FINDINGS and CONCLUSION.

\vspace{0.5em}

\textbf{User prompt template}\\
FINDINGS:
\{findings\}

CONCLUSION:
\{conclusion\}\\
Question: Is the conclusion consistent with the findings?\\
Answer with exactly one JSON object and no additional text:\\
\texttt{\{"consistent": true\}}
if the conclusion accurately reflects the findings and doesn't contain an error.\\
\texttt{\{"consistent": false\}}
if the conclusion contains an error, like contradicting the findings, misstating laterality (left/right), site (e.g. upper/lower, lung cancer instead of bone), or disease severity (e.g. faint uptake instead of extreme, normal instead of cancer).
\end{promptbox}

\textbf{P5 - constant $\rightarrow$ error}\\
Same as P4 above but instead of asking the models to classify the reports as consistent: true or false we ask them to classify the reports as error: true of false. 
\begin{promptbox}
\footnotesize
\textbf{System prompt}\\
You are a radiologist checking an oncology FDG PET/CT report for errors by comparing the FINDINGS and CONCLUSION.

\vspace{0.5em}

\textbf{User prompt template}\\
FINDINGS:
\{findings\}

CONCLUSION:
\{conclusion\}\\
Question: Is the conclusion consistent with the findings?\\
Answer with exactly one JSON object and no additional text:
\colorbox{red!10}{\texttt{\{"error": false\}}}
if the conclusion accurately reflects the findings and doesn't contain an error.\\
\colorbox{red!10}{\texttt{\{"error": true\}}}
if the conclusion contains an error, like contradicting the findings, misstating laterality (left/right), site (e.g. upper/lower, lung cancer instead of bone), or disease severity (e.g. faint uptake instead of extreme, normal instead of cancer).
\end{promptbox}

\textbf{P6 - Confidence prompt}\\
LLMs were prompted to return both their prediction and self reported confidence 1-5. We tried prompts both with and with the role of radiologist in the system prompt as we considered that this might be inflating model confidence but models still predicted mostly 5/5 confidence. 
\begin{promptbox}
\footnotesize
\textbf{System prompt}\\
You are a radiologist checking an oncology FDG PET/CT report for errors by comparing the FINDINGS and CONCLUSION.
\vspace{0.5em}

\textbf{User prompt template}\\
FINDINGS:
\{findings\}\\
CONCLUSION:
\{conclusion\}\\
Question: Is the conclusion consistent with the findings?\\
Answer with exactly one JSON object and no additional text:\\
\texttt{\{"consistent": true, "confidence": <integer 1-5>\}}  if the conclusion accurately reflects the findings and doesn't contain an error. \\
\texttt{\{"consistent": false, "confidence": <integer 1-5>\}} if the conclusion contains an error, like contradicting the findings, misstating laterality (left/right), site (e.g. upper/lower, lung cancer instead of bone), or disease severity (e.g. faint uptake instead of extreme, normal instead of cancer).
\vspace{-2mm}
\begin{tcolorbox}[
    colback=red!8,
    colframe=red!30,
    boxrule=0pt,
    arc=1mm,
    left=0mm,
    right=0mm,
    top=0mm,
    bottom=0mm
]
"confidence" is how strongly the error judgment is supported by the evidence (1 = not well supported, 5 = very well supported). When evidence is ambiguous, use confidence 1–2. When there is explicit evidence for your decision then use confidence 4-5.
\end{tcolorbox}
\end{promptbox}

\textbf{P10 - Few shot prompt}\\
We included 3 reports examples - 1 consistent (no-error) and 2 inconsistent (different errors) to P9 shown in Fig~\ref{fig:prompt}. P11 added 3 more to ensure one example of each kind of error featured in the prompt. The multi-class few-shot prompt was 6-shot (same as P11) - each error type labelled. See code for all prompts.

\begin{promptbox}
\footnotesize
\textbf{System prompt}\\
You are a radiologist checking an oncology FDG PET/CT report for errors by comparing the FINDINGS and CONCLUSION.
\vspace{0.5em}

\textbf{User prompt template}\\
\vspace{-5mm}
\begin{tcolorbox}[
    colback=red!8,
    colframe=red!30,
    boxrule=0pt,
    arc=1mm,
    left=0mm,
    right=0mm,
    top=0mm,
    bottom=0mm
]
Here are labeled examples of findings/conclusion consistency checks.

Example 1:\\
FINDINGS:
1. There is a 36mm, extremely FDG-avid (SUVmax=20.6) lobulated left upper lobe mass. No FDG avid lymphadenopathy. The remainder of FDG uptake is physiological. 2. Retro-oesophageal anomalous right subclavian artery. Large hiatus hernia containing the whole stomach. The remainder of the CT adds no further clinically relevant information. No evidence of brain metastases on the contrast-enhanced CT.\\
CONCLUSION:
Extremely FDG-avid right upper lobe mass, in keeping with a bronchogenic malignancy, if proven, stage T2a N0 M0.
Answer: \{"consistent": false\}

Example 2:\\
FINDINGS:
1. There is reduction in FDG avidity within the surgical bed in the medial aspect of the left thigh, SUV max 2.3 (previously 8.2), in keeping with post surgical inflammation. 2. There is linear FDG uptake within the muscle insertion into the left hip, proximal femur and right greater trochanter which is presumed inflammatory. 3. The remainder of FDG distribution is physiological. 4. Unenhanced CT component adds no other relevant findings.\\
CONCLUSION:
No FDG avid recurrent or residual disease demonstrated.
Answer: \{"consistent": true\}

Example 3:\\
FINDINGS:
1. The locally infiltrative lesion within the head of the pancreas is ill-defined and better demonstrated on recent contrast enhanced CT. It measures approximately 2.5 x 2.1 x 3.8 cm and is mildly FDG avid (SUVmax = 5.0). 2. No avid lymph nodes or metastatic deposits demonstrated. 3. The remaining FDG uptake is physiological. 4. Right lung subpleural nodules better characterised on the prior CT. The unenhanced CT component add no additional clinical information.\\
CONCLUSION:
Mildly FDG avid (SUVmax = 5.0) pancreatic lesion. No avid lymph nodes or definite metastases demonstrated. The marked FDG uptake in the primary tumour suggests that sensitivity for further disease is reduced.
Answer: \{"consistent": false\}

Now evaluate this report.
\end{tcolorbox}
\vspace{-1mm}
FINDINGS:
\{findings\}\\
CONCLUSION:
\{conclusion\}\\
Question: Is the conclusion consistent with the findings?\\
Answer with exactly one JSON object and no additional text:\\
\texttt{\{"consistent": true\}}  if the conclusion accurately reflects the findings and doesn't contain an error. \\
\texttt{\{"consistent": false\}} if the conclusion contains an error, like contradicting the findings, misstating laterality (left/right), site (e.g. upper/lower, lung cancer instead of bone), or disease severity (e.g. faint uptake instead of extreme, normal instead of cancer).\\
Minimize false alarms: only output \{"consistent\": false\} when an error is clearly and explicitly supported by the text. When uncertain, output \{"consistent": true\}.
\end{promptbox}

\textbf{Multi-class prompt}\\
We provided explicit definitions of each error type in the multi-class zero-shot prompt below. 
\begin{promptbox}
\footnotesize
\textbf{System prompt}\\
You are a radiologist checking an oncology FDG PET/CT report for errors by comparing the FINDINGS and CONCLUSION. Classify the error type using the standard multiclass taxonomy.
\vspace{0.5em}

\textbf{User prompt template}\\
FINDINGS:
\{findings\}

CONCLUSION:
\{conclusion\}\\
Task: Classify the relationship between the FINDINGS and CONCLUSION. Choose exactly one error type.\\

Error types (use these exact strings):\\
- consistent: findings and conclusion agree; no clinically meaningful mismatch or error.\\
- laterality\_swap: conclusion misstates left/right laterality relative to findings\\
- site\_swap: conclusion misstates spatial or anatomical site (e.g. lung vs liver, upper vs lower, pleural vs parenchymal)\\
- severity\_flip: conclusion misstates disease severity or uptake intensity (e.g. faint vs intense, mild vs severe, small vs large tumour)\\
- NF\_v\_Dis\_swap: conclusion contradicts whether findings show disease (no-findings vs disease conclusion swapped)\\
- cross\_pair\_mismatch: conclusion contradicts findings but the error is not laterality\_swap, site\_swap, severity\_flip, or NF\_v\_Dis\_swap (ie the whole conclusion has been swapped)\\

Answer with exactly one JSON object and no additional text:\\
\texttt{\{\{"error\_type": "$<$one of: consistent, laterality\_swap, site\_swap, severity\_flip, NF\_v\_Dis\_swap, cross\_pair\_mismatch$>$"\}\}}
\end{promptbox}

\subsection{Experimental details}

The dataset was randomly partitioned into training, validation, and test sets in an approximate 80:10:10 ratio. To prevent data leakage, the split was performed at the patient level, ensuring that all reports from a given patient were assigned to a single partition and that no patient's reports appeared in more than one split.
1,778 reports were excluded that didn't contain both findings and conclusions (216 were missing both) leaving us with 30,633 reports from 32,627. 

All experiments were conducted on a single NVIDIA RTX PRO 6000 GPU with 96\,GB of memory. 
We trained the \emph{domain-specific} tokeniser using a minimum frequency threshold of 3, such that only words or subwords appearing at least three times in the corpus were retained as tokens resulting in a vocabulary size of approximately 10k tokens \citep{Warr_2026}.
All domain-specific models share a 600-token context, GELU activations, and 0.1 hidden/attention dropout. 

\textbf{MLM pretraining} Encoder weights were first pretrained with masked language modelling (15\% token masking - off those 80\% masked, 10\% randomly replaced, 10\% replaced with the original token) for 50 epochs, batch size 32, AdamW with weight decay 0.01 and a linear learning-rate decay. The 4- and 6-layer models used peak learning rate $5\times10^{-4}$ with no warmup (fp16); the 8- and 12-layer models used 
$2\times10^{-4}$ with 1,000 warmup steps (bf16).

\textbf{CLS pretraining}
A findings–conclusion consistency head was then trained for 20 epochs with AdamW at $2\times10^{-5}$ and batch size 128 (64 for the 12-layer model), using the same optimiser settings as the downstream error-detection stage (constant learning rate, no warmup).

\begin{figure}[h]
    \centering
    \includegraphics[width=\linewidth]{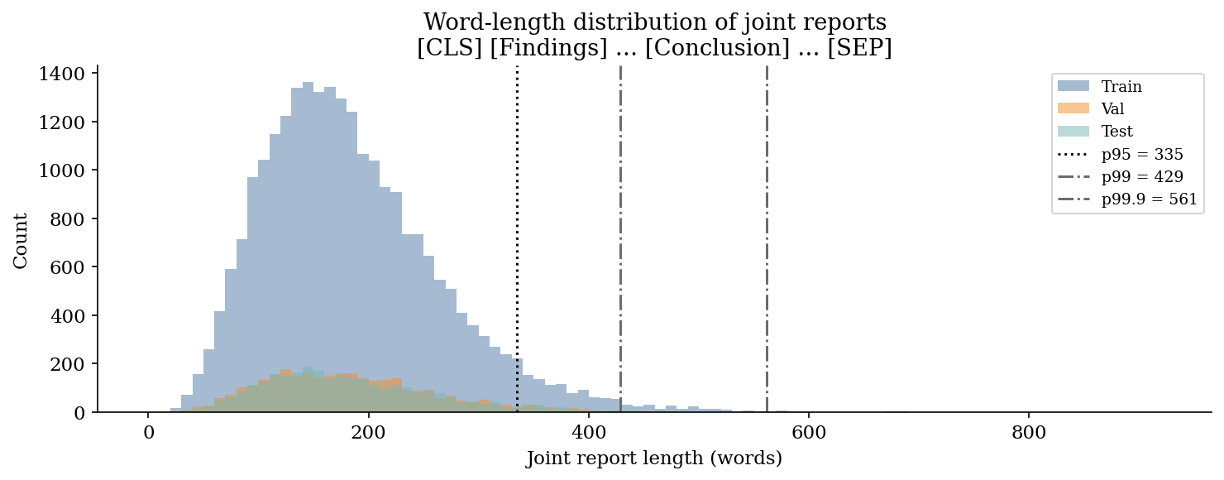}
    \caption{The length distribution of the joint findings and conclusion with 95, 99th and 99.9th percentiles shown.}
    \label{fig:len_words}
\end{figure}

Fig.~\ref{fig:len_words} show the distribution of the length (number of words) in the PET/CT reports. The input to the LLMs is slightly longer than domain specific due to both the natural language tokeniser and the need to include a prompt.

\section{Appendix: Additional Results}

\subsection{Model performance breakdown}
Tab.~\ref{tab:results} show binary results from main paper (Tab.~2) broken down by category (5 errors and no error/ consistent reports). The LLM results are from the best prompt. The best prompt was defined as the prompt that gave the best binary balanced accuracy - an average of macro recall of all the 5 error types and the true negative rate (TNR - recall of the no error class). This was chosen to reflect that real reporting errors do not make up the majority of cases and we at least wanted to give equal weighting to the consistent class as high false positive rates (FPR) mean practical deployment for these models would be impossible.

\begin{table}[h]
    \centering
    \caption{Binary error detection rate, broken down by error type. For best LLM prompt.}
    \renewcommand{\arraystretch}{1}
    \setlength{\tabcolsep}{5pt}
    \resizebox{\textwidth}{!}{
    \begin{tabular}{lcccccc}
        \toprule
        \textbf{Model} &
        \textbf{Conclusion swap} &
        \textbf{Disease status} &
        \textbf{Disease severity} &
        \textbf{Laterality} &
        \textbf{Anatomical site} &
        \textbf{No Error} \\
        \midrule
        Qwen3      & 95.6 & 76.2 & 61.0 & 76.4 & 63.6 & 91.7 \\
        Gemma3     & 97.0 & 92.0 & 60.8 & 84.6 & 74.4 & 81.9 \\
        Llama3     & 96.0 & 81.6 & 69.8 & 90.2 & 73.0 & 85.8 \\
        \midrule
        BERT$_4$   & 90.6 & 97.1 & 71.3 & 11.0 & 34.8 & 90.9 \\
        BERT$_6$   & 97.6 & 97.5 & 92.1 & 94.4 & 91.3 & 94.2 \\
        BERT$_8$   & 98.2 & 99.6 & 94.8 & 96.2 & 92.1 & 90.7 \\
        BERT$_{12}$& 99.0 & 99.0 & 96.6 & 96.0 & 95.3 & 92.1 \\
        \bottomrule
    \end{tabular}
    }
    \label{tab:results}
\end{table}

\subsection{Fine-tuning Llama}
We adapted the best performing generic LLM (Llama) with the best performing prompt (P8) to the task.
We fine-tuned Llama-3.3-70B-Instruct with QLoRA~\citep{Dettmers_qlora_2023, lora_2021}, freezing a 4-bit NF4-quantized base model and training LoRA adapters (r=16, $\alpha$=32, dropout=0.05) on all attention and feed-forward projections for one epoch of supervised fine-tuning. We used the same training set as the domain-specific models: 24,525 findings–conclusion consistency examples, with errors generated in the same way as domain-specific training. We used an AdamW optimiser with learning rate $2\times10^{-4}$, cosine decay with 3\% warmup, gradient accumulation to an effective batch size of 8 (from batch size of 2), and a maximum sequence length of 2048.
Results are shown in Tab.~\ref{tab:fine-tune}. 
Fine-tuning closed the gap in performance but didn't surpass the compact domain-specific models despite its LoRA adapter alone containing approximately 200M parameters, more than twice the 94M parameters of the largest compact domain-specific model. Fine-tuned LLM inference required over 4hrs, compared with 30–60s for the domain-specific models, representing a >150-fold increase in inference time.

\begin{table}[t]
    \centering
    \caption{Performance and computational requirements of generic LLMs, QLoRA-adapted Llama-3.3-70B, and compact domain-specific models on the held-out test set. Fine-tuning closed the performance gap between generic LLMs and compact domain-specific models but did not surpass the compact models, while requiring substantially more inference parameters and time.}
    \resizebox{\linewidth}{!}{
    \begin{tabular}{llc ccc}
    \toprule
    \textbf{Model} & \textbf{Task adaption}&\textbf{Trainable param.} &\textbf{Inference param.} &\textbf{Bal. Acc.}& \textbf{Inference time}\\
        \midrule
        Gemma3 &  None & 0 & 27B & 81.8 & $\sim$5hrs45 \\
        Qwen3 &  None & 0 & 32B & 83.3 &  $\sim$5hrs15\\
        Llama3 &  None & 0 & 70B & 84.0 & $\sim$2hrs10\\
        \midrule
        Llama3-QLORA  &  Partial & 207M & 70B & 94.4 & $\sim$4hrs20 \\
        \midrule
        BERT$_6$ &  Full & 15M & 15M & 94.4 & $\sim$30s\\
        BERT$_{12}$ &  Full & 94M & 94M & 94.6 & $\sim$1min \\
    \bottomrule
    \end{tabular}}
    \label{tab:fine-tune}
\end{table}

\subsection{Multi-class results}
\begin{figure}[h]
    \centering
    \includegraphics[width=0.75\linewidth]{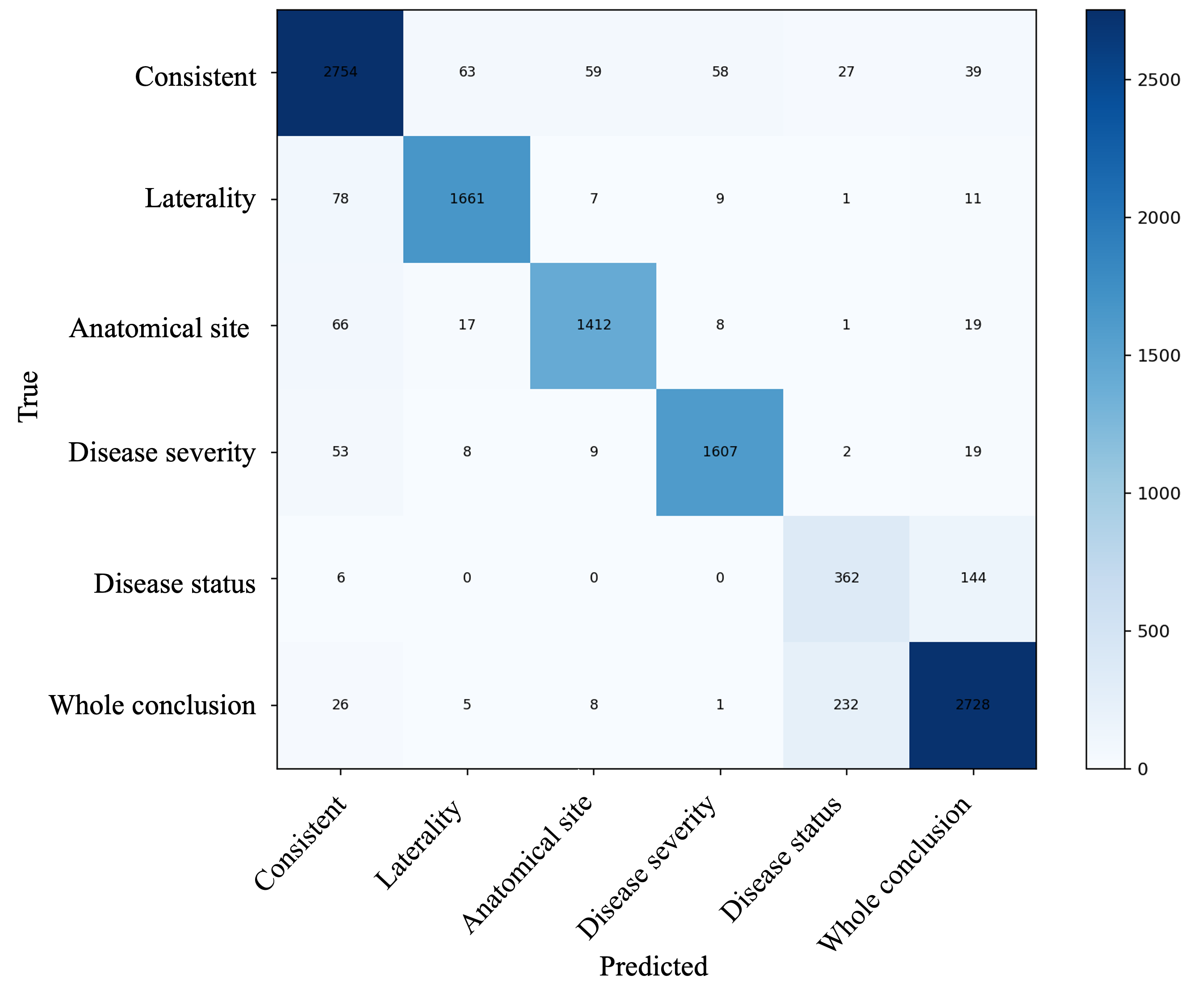}
    \caption{BERT$_{12}$ multi-class confusion matrix.}
    \label{fig:mc_cm}
\end{figure}

The multi-class confusion matrix for BERT$_12$ is shown in  Fig.~\ref{fig:mc_cm}. 
A high source of incorrectly classified reports for the domain-specifc models comes from misclassification of categories of whole conclusion swap and no findings vs disease swap -  categories which do overlap. 

\subsection{Domain-specific BERT held out error}
\begin{table}[h]
\centering
\caption{Severity term-holdout generalisation on the test benchmark (binary BERT$_6$). Each row reports severity recall of the models and the held-out terms for each model.   
Remaining severity is the complementary subset of \emph{severity error} rows whose swapped term was not withheld.}
\label{tab:severity-term-holdout}
\small
\resizebox{\linewidth}{!}{
\begin{tabular}{l ccccccc}
\toprule
Holdout terms & $n$ & Baseline Model & Holdout Model & Baseline Terms & Held-out Terms & $\Delta$ & Remaining \\
\midrule
mild(ly) / severe(ly) 
& 417 & 92.1\%  & 77.3\% & 96.2\% & 71.7\% & $-24.5$ & 78.9\% \\
mild(ly) / intense(ly)
& 483 & 92.1\% &77.4\% & 96.3\% & 80.1\% & $-16.2$ & 76.7\% \\
faint(ly) / intense(ly) 
& 136 & 92.1\%  & 92.0\%& 97.1\%&95.6\% & $-1.5$ & 91.9\%  \\
\bottomrule
\end{tabular}}
\end{table}

\begin{table}[h]
\centering
\caption{Severity term-holdout generalisation on the test benchmark (binary BERT$_{12}$). Each row evaluates the same held-out term slice under the full-severity baseline and the corresponding holdout-trained model. Remaining severity is the complementary subset of \emph{severity error} rows whose swapped term was not withheld.}
\label{tab:severity-term-holdout}
\small
\resizebox{\linewidth}{!}{
\begin{tabular}{l ccccccc}
\toprule
Holdout terms & $n$ & Baseline Model & Holdout Model& Baseline Terms & Holdout Terms & $\Delta$ & Remaining \\
\midrule
mild(ly) / severe(ly)
  & 417 & 96.6\% &  79.0\% & 97.4\% & 71.5\% & $-25.9$ & 81.1\% \\
mild(ly) / intense(ly)
  & 483 & 96.6\%&  84.0\% & 97.7\% & 85.7\% & $-12.0$ & 83.1\% \\
faint(ly) / intense(ly)
  & 136 & 96.6\% & 91.6\% & 100.0\% & 91.2\% & $-8.8$ & 91.6\% \\
\bottomrule
\end{tabular}}
\end{table}

\subsection{Calibration Analysis}
\label{append:cal}
\begin{figure}[h]
    \centering
    \includegraphics[width=\textwidth]{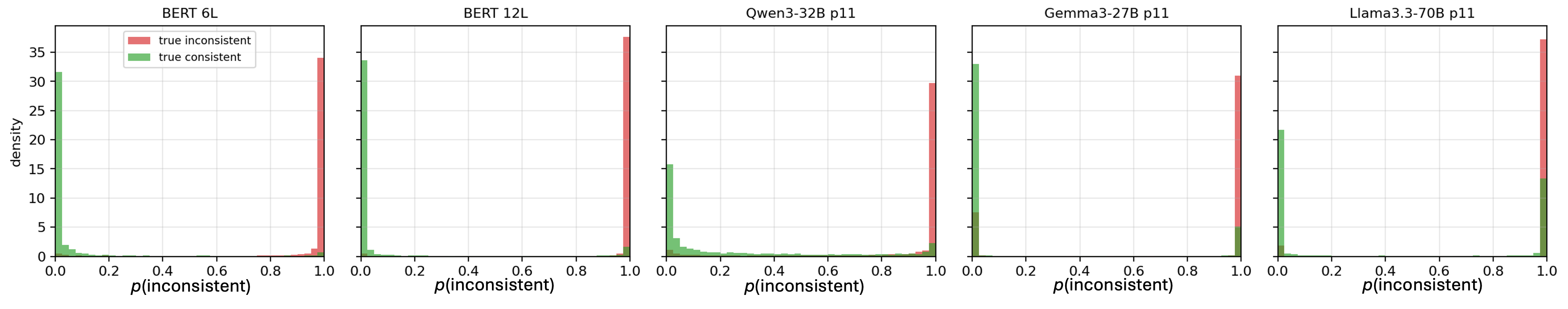}
    \caption{Probabilities of the models predicting whether or not a report contained an error coloured by ground truth label. Using confidence prompt (P6) for the LLMs to compare with self-reported confidence. Margin between predictions of error are close to one - i.e. all models are extremely confident.}
    \label{fig:tok_probs}
\end{figure}

We assessed calibration of the inconsistency score P(inconsistent) using expected calibration error (ECE; 10 equal-width bins) shown in Main Fig.~\ref{fig:ece_mce}. The generic-domain LLMs probabilities were taken from the top-k probabilities of the true or false token (k=5) and if false was in the top k words predicted that probability ($P(inconsistent)$) was stored. Scores are unnormalised JSON-boolean token probabilities, and $P(true)+P(false)\approx1$ so renormalizing does not change ECE \citep{Kadavath_2022}. For domain-specific binary BERTs $P(inconsistent)$ is just the probability of the prediction.
BERT models with $\geq$6 layers were well calibrated (ECE 0.027–0.033), whereas the 4-layer model was substantially worse (ECE 0.165). Among LLMs (prompt 11), Qwen3-32B was best calibrated (ECE 0.085), followed by Llama3.3-70B (0.137) and Gemma3-27B (0.185). Reliability diagrams (Main Fig.~\ref{fig:ece_mce}(c)) and Fig.~\ref{fig:tok_probs} show that scores are highly concentrated near 0 and 1 for all models except BERT-4L (39\% of predictions in the edge bins, vs 80–98\% for the others).
Consequently, ECE largely reflects whether extreme predictions are trustworthy: for the stronger BERT models, near-0 scores correspond to truly consistent pairs and near-1 scores to truly inconsistent pairs, whereas Gemma and Llama retain large absolute gaps in these dominant bins despite similar edge concentration. ECE was stable under alternative bin counts (5–50), with unchanged model ranking. 
We report calibration of the models' uncalibrated probabilities; post-hoc calibration methods such as temperature scaling were not explored here but may be done in future work \citep{Guo_2017}.

Plots of the probabilities of the model predicting error shown in Fig.~\ref{fig:tok_probs}.  All plots show the models confidence in predictions were high - predictions are concentrated at either end of the plots with very little in the middle (margins close to one). This indicates that the LLMs' self reported confidence (Main Fig.~\ref{fig:atn_conf}b) was at least aligned with the token (true/false) prediction.
BERT$_{12}$'s predictions are the most confident. 

\subsection{Analysis of BERT models decision influence}

\begin{figure}[h]
    \centering
    \includegraphics[width=\textwidth]{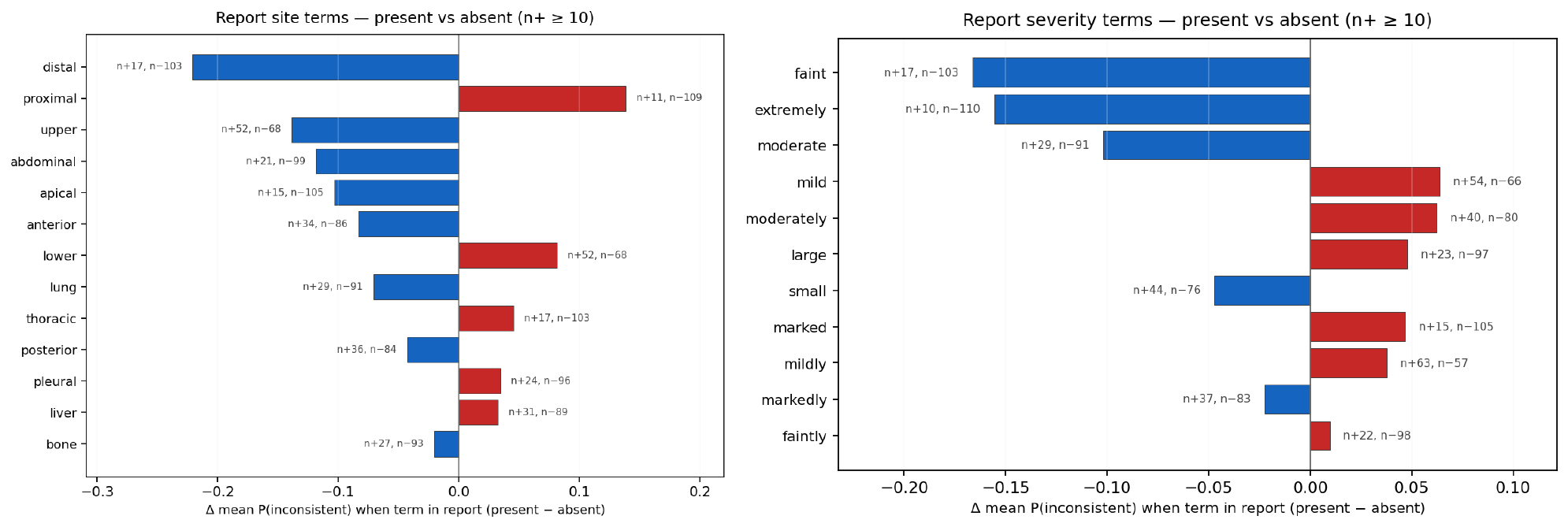}
    \caption{Association between anatomical site (left) and severity (right) terms and BERT$_{12}$ model-predicted inconsistency probability on the evaluation set. For each site keyword from the site-swap lexicon, bars show the difference in mean predicted probability of inconsistency between reports containing the term and reports not containing it (present - absent). Only terms appearing in at least 10 sampled reports are shown (n $+\geq$ 10); annotations give counts with (n$+$) and without (n$-$) the term. Red bars indicate higher predicted inconsistency when the term is present; blue bars indicate lower predicted inconsistency.
}
    \label{fig:site_lift}
\end{figure}

Fig.~\ref{fig:site_lift} show the influence of the key-word terms on models predicting error for each anatomical site and severity errors respectively.
To assess whether the classifier’s predictions co-vary with clinically meaningful lexical cues, we also computed key-word lifts over the evaluation corpus. For each predefined site or severity key-word, we compared the mean predicted probability of inconsistency for reports containing that term anywhere in the full report text (findings + conclusion) versus reports lacking it. The lift is defined as:
\begin{equation}
    \Delta \bar P = \mathbb{E}[P(inconsist.)|\text{term present}]-\mathbb{E}[P(inconsist.)|\text{term absent})] 
\end{equation}
Terms were drawn from the site-swap and severity-flip vocabularies used to construct synthetic inconsistencies during dataset generation. To avoid unstable estimates from rare terms, we retained only key-words with at least 10 present-case reports in the analysed sample (n$+\geq$ 10). This analysis reflects behavioural association between surface forms and model outputs; it is complementary to token-level attribution methods (e.g. attention scores in main paper Fig.~3a) and does not, by itself, establish causal importance.

\textbf{Site} Reports containing proximal, lower, or thoracic site terms received higher predicted inconsistency scores, whereas distal, upper, and abdominal terms were associated with lower scores, with paired antonyms often showing opposite effects.\\
\textbf{Severity} Severity descriptors were not uniformly ordinal predictors of inconsistency; mild/moderately/marked terms increased predicted inconsistency, while faint/extremely/moderate decreased it, and morphological variants of the same root could show opposite associations. 

\subsection{Error incidence plots}
\begin{figure}[h]
    \centering
    \includegraphics[width=\linewidth]{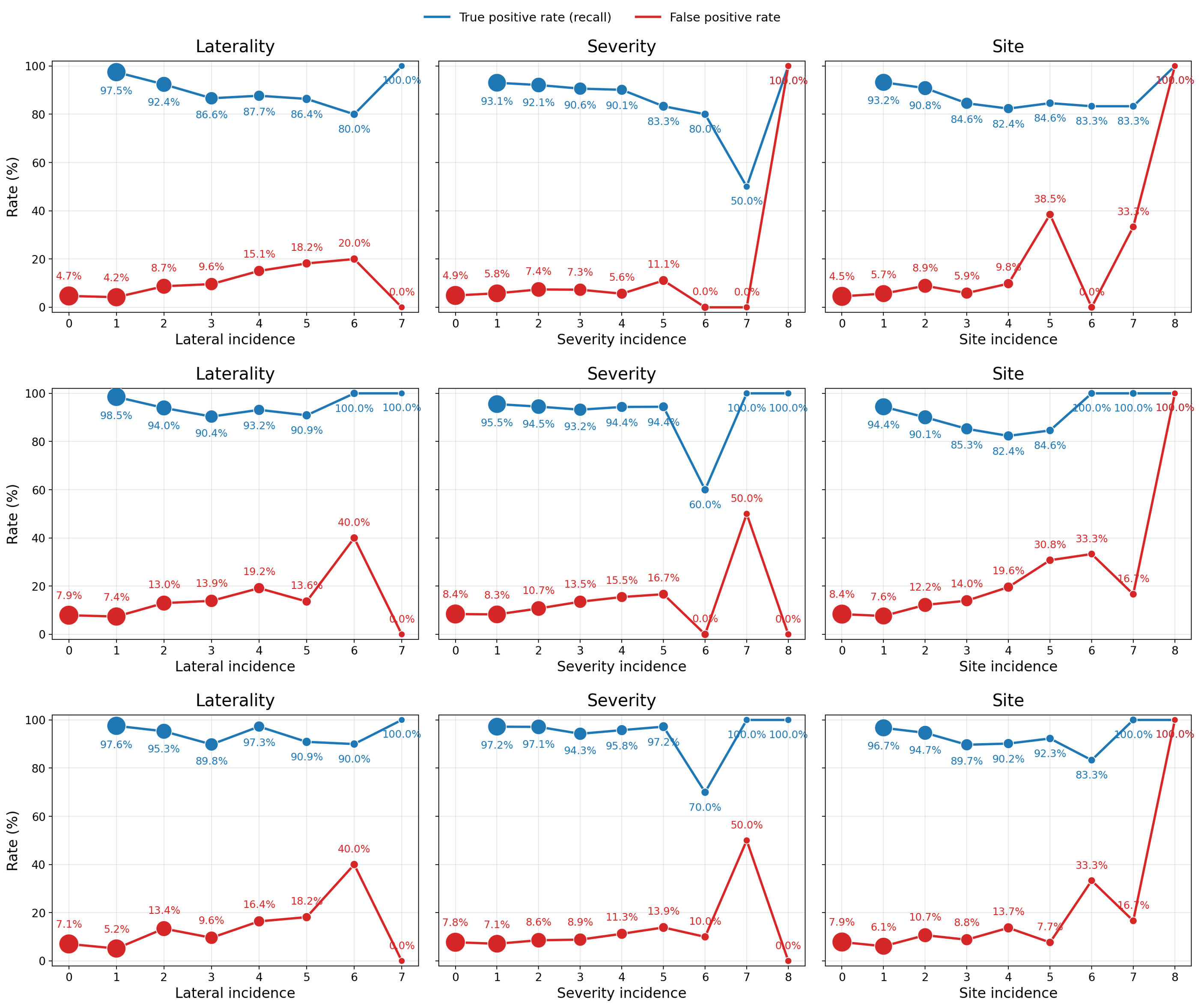}
    \caption{Domain-specific BERT TPR and FPR by incidence for the three different keyword errors (laterality, disease severity, anatomical site). (top) 6 layers, (middle) 8 layers, (bottom) 12 layers. Marker size = number of conclusions with that error incidence.}
    \label{fig:bert_inc}
\end{figure}

\begin{figure}[h]
    \centering
    \includegraphics[width=\linewidth]{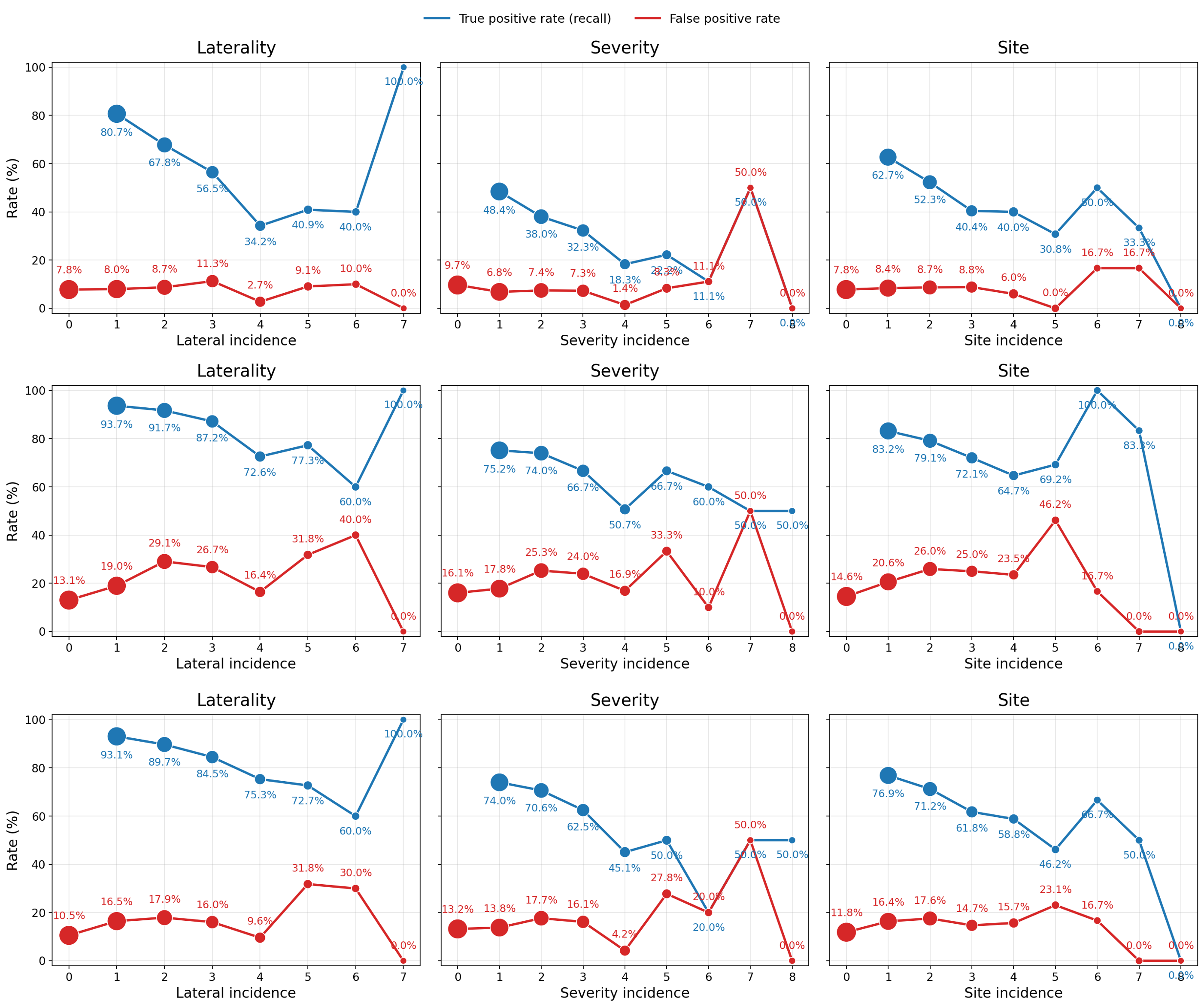}
    \caption{Generic-domain LLMs TPR and FPR by incidence for the three different keyword errors for P7. (top) Gemma, (middle) Qwen, (bottom) Llama}
    \label{fig:llm_inc}
\end{figure}

All erroneous reports contained just one incidence of an error that we deliberately inserted. However, for key-word swaps there may be multiple potential ways for us to insert an error (several incidences of left/right in the conclusion for example). 
We plotted detection rate and FPR for each of the keyword swaps (T$_{lat}$, T$_{anat}$, T$_{sev}$ ) to see if performance changed depending on the number of keywords present to potentially confuse the models. Fig.~\ref{fig:bert_inc}\ref{fig:llm_inc} show these plots for BERT and LLMs respectively. 
There is a small degradation in performance of the BERT models as keyword presence increases but a larger one for the LLMs particularly in detection rate.


\end{document}